\documentclass[journal]{IEEEtran}
\usepackage[T1]{fontenc}
\usepackage{graphicx}
\usepackage{times}
\usepackage{helvet}
\usepackage{booktabs}
\usepackage{courier}
\usepackage{amsmath}
\usepackage{algorithm}
\usepackage{algorithmic}
\usepackage{csquotes} 
\usepackage{color}
\usepackage{paralist}
\usepackage{amssymb}
\usepackage{indentfirst}
\usepackage{subfigure}
\usepackage{float}
\usepackage{multirow}
\usepackage{cite}
\usepackage{mathrsfs}
\usepackage[colorlinks, linkcolor=red,  anchorcolor=blue, citecolor=blue]{hyperref}
\usepackage{textcomp,booktabs}
\usepackage{amssymb}
\usepackage{pifont}
\newcommand{\cmark}{\ding{51}}%

\begin{document}

\title{ Tracking by Joint Local and Global Search: A Target-aware Attention based Approach } 

\author{Xiao Wang, Jin Tang, Bin Luo,  \emph{Member, IEEE}, Yaowei Wang, \emph{Member, IEEE}, Yonghong Tian, \emph{Senior Member, IEEE}, Feng Wu, \emph{Fellow, IEEE} 

\thanks{Xiao Wang, Yaowei Wang, Yonghong Tian, and Feng Wu are with Peng Cheng Laboratory, Shenzhen, China. Jin Tang and Bin Luo are with the School of Computer Science and Technology, Anhui University, Hefei, China. Professor Feng Wu is also from University of Science and Technology of China, Hefei, China. Yaowei Wang, Yonghong Tian are also with the National Engineering Laboratory for Video Technology, School of Electronics Engineering and Computer Science, Peking University, Beijing, China. 

Corresponding author: Feng Wu and Yaowei Wang. 
Email: \{wangx03, tianyh, wangyw\}@pcl.ac.cn, \{tangjin, luobin\}@ahu.edu.cn, fengwu@ustc.edu.cn}}

\markboth{IEEE Transactions on Neural Networks and Learning Systems} 
{Shell \MakeLowercase{\textit{et al.}}: Bare Demo of IEEEtran.cls for IEEE Journals}

\maketitle

\begin{abstract}	
Tracking-by-detection is a very popular framework for single object tracking which attempts to search the target object within a local search window for each frame. Although such local search mechanism works well on simple videos, however, it makes the trackers sensitive to extremely challenging scenarios, such as heavy occlusion and fast motion. In this paper, we propose a novel and general target-aware attention mechanism (termed TANet) and integrate it with tracking-by-detection framework to conduct joint local and global search for robust tracking. Specifically, we extract the features of target object patch and continuous video frames, then we concatenate and feed them into a decoder network to generate target-aware global attention maps. More importantly, we resort to adversarial training for better attention prediction. The appearance and motion discriminator networks are designed to ensure its consistency in spatial and temporal views. In the tracking procedure, we integrate the target-aware attention with multiple trackers by exploring candidate search regions for robust tracking. Extensive experiments on both short-term and long-term tracking benchmark datasets all validated the effectiveness of our algorithm. The project page of this paper can be found at \url{https://sites.google.com/view/globalattentiontracking/home/extend}.  
\end{abstract}


\IEEEpeerreviewmaketitle

\section{Introduction}
\IEEEPARstart{V}{isual} tracking is to estimate the states of target object in sequential video frames, given the initial bounding box. It is one of the most important research topics in the computer vision community and has many practical applications such as autonomous driving, visual surveillance, and robotic. Many trackers are proposed in recent years, but visual tracking still faces many challenges including heavy occlusion, abrupt changing, and large deformation, etc. How to achieve robust and accurate tracking in challenging scenarios is still an open problem. 

\begin{figure*}[!htb]
\center
\includegraphics[width=7in]{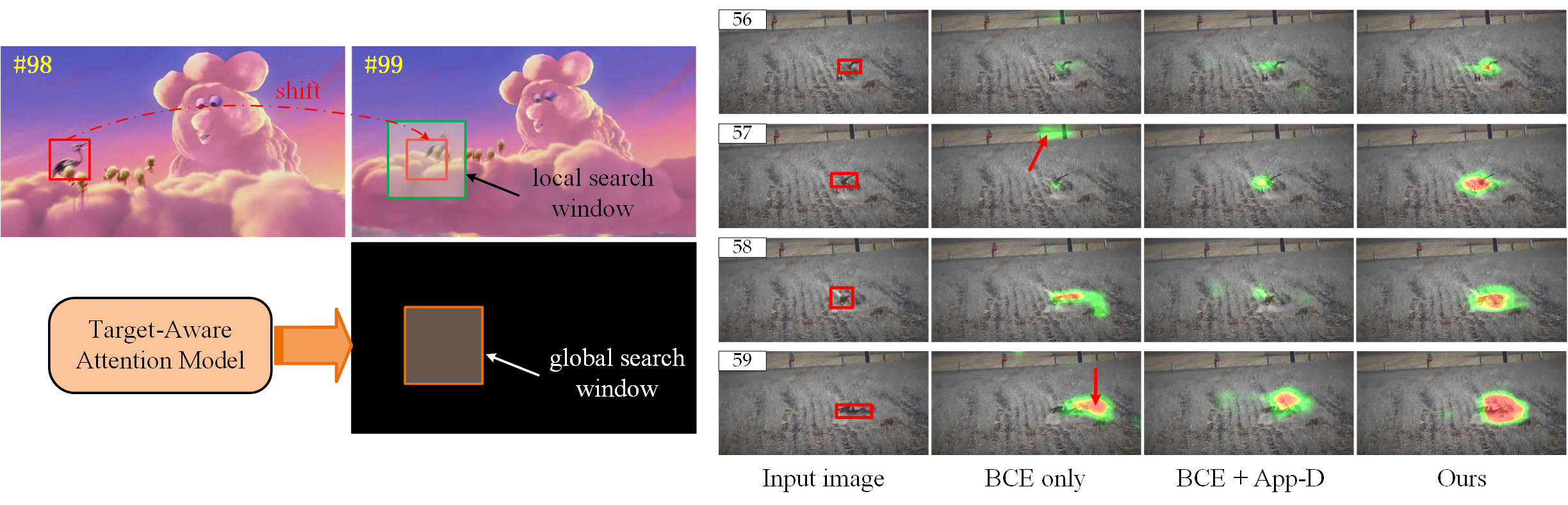}
\caption{The left sub-figure is our proposed tracking by joint local and global search, the right sub-figure is used to compare the attention maps predicted with different models.}
\label{motivation}
\end{figure*}

As is known to all, many trackers are developed based on the \emph{tracking-by-detection framework} \cite{li2018siamRPN, song2017crest, Nam2015Learning, Danelljan2016ECO, Danelljan2016Beyond, hu2017manifold, zhu2018dynamic, ge2020cascaded, yun2018action, liwicki2012efficient, wang2021tnl2k} and achieve good results on simple videos. However, their performance is still unsatisfying in challenging scenarios. Take this question in mind, we begin to revisit the tracking-by-detection framework and find the following procedures may be the key issues that lead to model drift. 
Firstly, they simply set a local search window using temporal information. The validation of this assumption is that the motion of the target object between continuous two frames is relatively small. We think this may not hold in practical tracking scenarios, especially when fast motion, heavy occlusion, deformation occurred. Therefore, these trackers always lose the target in extremely challenging frames. And it is difficult to re-track the target since the searching window is already invalid for sampling target candidates. 
Secondly, they first estimate the target location using a fixed scale and then determine the scale of the target at the estimated center location. Such approaches heavily rely on the object location and their performance could be easily disturbed when the location is unreliable. Moreover, fewer sampled scales may exclude the true state of the target, while more sampled scales will introduce high computational cost. Such local search strategy works well on simple videos, however, they still suffer from the aforementioned challenging factors, even deeper and wider networks adopted \cite{li2018siamRPN, zhang2019deeper, li2019siamrpn++}. Surprisingly, seldom of previous works argue the local search strategy of the tracking-by-detection framework.

The aforementioned analysis inspires us to begin to think can we design a novel search scheme for visual tracking? One intuitive approach is to conduct global search with RPN-based method \cite{huang2019globaltrack}, however, only global search based trackers will be easily influenced by similar objects. Another idea is to jointly utilize the local and global search for tracking as \cite{zhu2016beyond, zhang2018learningTermTracking, yan2019skimming} does. However, the edge-based feature used in  \cite{zhu2016beyond} is still unreliable when the background is clutter which further limits their tracking performance. The authors of \cite{zhang2018learningTermTracking, yan2019skimming} attempt to conduct global tracking in a sliding window manner which will be time-consuming. Therefore, how to mine the candidate search regions which most related to the target object is the key insight for joint local and global search strategy.

\begin{figure*}[!htb]
\center
\includegraphics[width=6in]{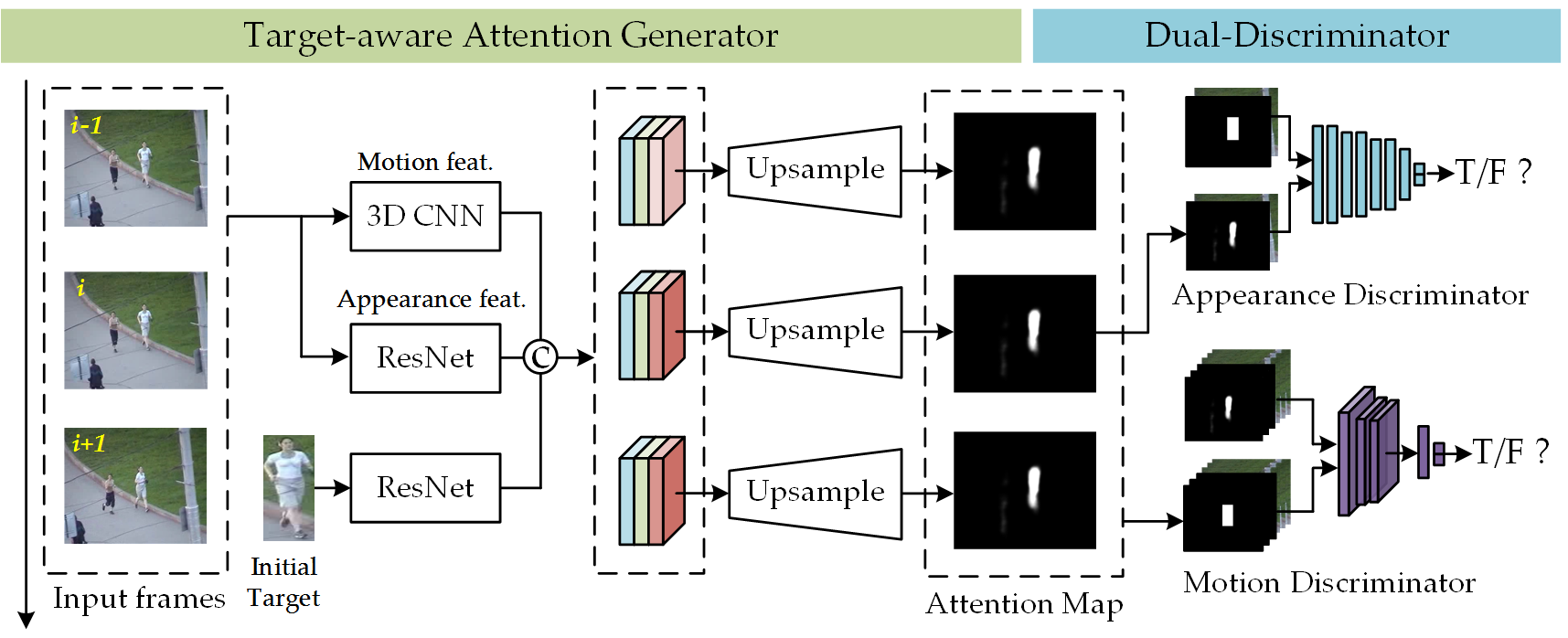}
\caption{The pipeline of our proposed target driven attention network for joint local and global search for visual object tracking. The BCE loss is ignored for the simplicity. }
\label{pipeline}
\end{figure*}

Inspired by the recent progress of semantic segmentation \cite{long2015fullyFCNsegmentation, chen2018deeplab}, in this paper, we propose a novel and general target-aware attention learning approach to achieve global search under the tracking-by-detection framework. As shown in Fig. \ref{pipeline}, our Target-aware Attention Network (named as TANet) takes several consecutive video frames and the target object as inputs and generates corresponding attention maps as output. In particular, we input consecutive frames as the input of 3D CNN to capture motion information, and current frame into 2D CNN to learn the appearance information. More importantly, we integrate the initialized target object in the first frame into the learning framework for target-specific feature learning. These features will be concatenated and fed into a decoder network to generate corresponding attention maps. The proposed TANet can be trained with \emph{pixel-wise classification loss} which is widely used in segmentation related tasks \cite{gong2017lookintoPerson, liang2015deepParse}. Although it can provide us a good result, however, it is designed for per-pixel category prediction, as noted in \cite{luo2018macroAdversarial}, and it has the following two shortcomings: 
1). the pixel-wise classification loss function may lead to local inconsistency, because it merely penalizes the false prediction on every pixel without explicitly modeling the correlation among adjacent pixels; 
2). the pixel-wise classification may lead to semantic inconsistency in the global attention map. As shown in Fig. \ref{motivation} (right sub-figure), the attention maps generated by binary cross-entropy loss function only are easier affected by background clutter.

To handle the inconsistency problems, a recent works resort to adversarial network \cite{luc2016semanticGAN, tang2019salientDetection} due to the adversarial loss judge whether a given attention map is real or fake by joint configuration of many label variables, and enforce higher-level consistency. And many works adopt the routine of combining the cross-entropy loss with an adversarial loss to produce the label maps closer to the ground truth \cite{wang2019qualityRGBTsaliency, luo2018macroAdversarial, nguyen2017shadowDetectionGAN, wang2018stackedDetectionGAN}. In our conference paper \cite{wang2019GANTrack}, we follow this setting to train our generator and get better attention maps for robust tracking. Meanwhile, we find that the predicted attention map is not always consistent between continuous video frames. Such inconsistent attention will bring us \emph{jumping} candidate proposals which may hurt the tracking performance. In this paper, we intuitively design a motion discriminator to address the inconsistency issue by conducting adversarial training in the motion view. Experimental results demonstrate that our proposed TANet can significantly boost the tracking results of multiple trackers on multiple benchmark datasets.

Generally speaking, the highlights of our work can be concluded as follows: \\ 
- \textbf{Flexible:} Our proposed target-aware attention is a flexible global search module for single object tracking. As a plug and play module, it can be integrated with many existing trackers as validated in our experiments. \\ 
- \textbf{Effective:} In our experiments, we conduct extensive experiments on six large-scale tracking datasets, including short-term and long-term datasets. These results demonstrate that our attention is effective for both short-term and long-term tracking tasks. \\ 
- \textbf{Efficient:} Our attention model only costs 0.18 seconds for each image and can be used selectively when encountering challenging factors. For instance, when integrating TANet with SiamRPN++, it still can running at 35 and 134 FPS if the ResNet-50 and AlexNet are used as the backbone, respectively. Therefore, the overall efficiency of tracking can be ensured, meanwhile the overall performance can be significantly improved.

To sum up, we draw the contributions of this paper as follows: 

$\bullet$ We analyze the limitations of local search policy in popular tracking-by-detection framework and propose the \emph{joint local and global search framework} to handle these issues.

$\bullet$ We propose a simple yet effective Target-aware Attention Network (TANet) and design a novel adversarial learning scheme for the training of our generator. In more detail, we use appearance and motion discriminator to ensure the spatial and temporal consistency of predicted target-aware attention maps.

$\bullet$ We demonstrate that our proposed TANet is a plug and play module by integrating it with multiple baseline trackers on several popular tracking benchmark datasets. Extensive experiments on both short-term and long-term tracking datasets validated the effectiveness and generalization of our proposed algorithm.

This paper is an extension of our conference paper \cite{wang2019GANTrack}. The main difference between this paper and the conference version is concluded as follows: 
1). We re-design the generator network to predict the attention maps of continuous frames simultaneously in the training phase; 
2). We design a novel motion discriminator to augment the target-aware attention generation from the perspective of temporal view; 
3). We validate the effectiveness and generalization of our model on both short-term and long-term large-scale tracking benchmark datasets.

\section{Related Work}

In this section, we will give a brief review of the local and global search based trackers, visual attention based trackers, and generative adversarial learning. Due to the limited space in this paper, more related works on single object trackers can be found in the following survey papers \cite{li2018TrackSurvey, marvasti2021TrackSurvey, brunetti2018CVDLSurvey} and paper list\footnote{\url{https://github.com/wangxiao5791509/Single_Object_Tracking_Paper_List}}.

\textbf{Local Search based Tracking: } 
For short-term tracking, existing trackers usually adopt local search strategy for target localization. These tracking algorithms can be divided into three main streams, i.e., binary classification based trackers \cite{Nam2015Learning, han2017branchout, Park_2018_ECCV, jung2018real, SongYiBing_2018_CVPR, Li2017ReGLe, hare2015struck}, Correlation Filter (CF) based trackers \cite{danelljan2015learning, choi2017attentional, Bolme2010Visual, valmadre2017end, lukezic2017discriminative, Danelljan2016Beyond, Henriques2015High}, and Siamese network based trackers. Specifically, the binary classification based trackers treat visual tracking as a classification problem and discriminate the given proposal is the target or not with an online learned model. Usually, the support vector machine (SVM) \cite{wang2005svm} or deep neural networks (DNNs) \cite{lecun2015DLnature} is adopted to achieve this goal. In the deep learning era, the DNNs based classification usually attains better results than SVM based trackers. Representative trackers like MDNet  \cite{Nam2015Learning} is designed to pre-train a CNN using a large set of videos with tracking ground truth to obtain a generic target representation. Many works are developed based on this idea, such as the ensemble learning based BranchOut \cite{han2017branchout}, meta-learning based tracker  \cite{Park_2018_ECCV}, and real-time MDNet \cite{jung2018real}. Although these works achieve higher performance, however, the running efficiency and model complexity limit their applications. Some deep learning based CF trackers \cite{danelljan2015learning, choi2017attentional, valmadre2017end, hu2017manifold} can achieve more efficient tracking, however, their performance is outperformed by Siamese trackers by a large margin, especially on the large-scale benchmark datasets like LaSOT \cite{fan2019lasot}, TrackingNet \cite{muller2018trackingnet}.

In recent five years, the Siamese trackers occupy the vast majority of algorithms in the tracking field. The pioneering works like SiamFC \cite{bertinetto2016fully} and SINT \cite{Tao2016Siamese} first exploit the response maps and matching scores between target template and local search region with Siamese network, respectively. GOTURN \cite{held2016goturn} also draws more and more attention from researchers due to its high efficiency. After that, the regional proposal networks is introduced (i.e., SiamRPN \cite{li2018siamRPN}, SiamRPN++ \cite{li2019siamrpn++}) and deeper and wider backbone networks are design for high-performance tracking \cite{zhang2019siamDW, xu2020siamfc++}. The adversarial training is adopted to improve the robustness of trackers \cite{Wang_2018_CVPR, SongYiBing_2018_CVPR, zhao2018adversarial}. The ATOM \cite{danelljan2019atom}, DiMP \cite{Bhat_dimp}, prDiMP \cite{danelljan2020prDIMP} and D3S \cite{lukezic2020d3s} also demonstrate the effectiveness of discriminative feature learning for tracking. 
Due to the utilization of local search, many of these trackers still sensitive to challenging factors including \emph{out-of-view}, \emph{heavy occlusion}, and \emph{fast motion}. Therefore, many global search based trackers are developed to address the aforementioned issues which can be found in the subsequent paragraph.

\textbf{Global Search based Tracking: } 
To address aforementioned issues, many researchers begin to design global search based trackers \cite{fang2019LGTrack, ma2015LTCF, ramesh2018LTEvent, zhu2016beyond, huang2019globaltrack, voigtlaender2019mots, zhang2020TRL, zhang2018learningTermTracking, yan2019skimming, xuan2021siamLT, dai2020high, voigtlaender2020siamRCNN}. Specifically, Ma \emph{et al.} \cite{ma2015LTCF} train an online random fern classifier to re-detect objects in case of tracking failure. The authors of \cite{ramesh2018LTEvent} also adopt local-global search for event camera based tracking by switching between local tracking and global detection. Zhu \emph{et al.} \cite{zhu2016beyond} utilize hand-designed features for global proposal generation which are too simple for practical tracking scenarios. Huang \emph{et al.} propose to tracking by global search only \cite{huang2019globaltrack} and achieve good performance on long-term tracking datasets. However, their model performs poorly in short-term benchmarks due to challenging factors like similar objects. Paul \emph{et al.} \cite{voigtlaender2019mots} propose a baseline method that jointly addresses detection, tracking, and segmentation with a single CNN model. DTNet is proposed in \cite{zhang2020TRL} by Song \emph{et al.} to integrate object detection and tracking into an ensemble framework based on hierarchical reinforcement learning. Some trackers conduct the global search in a slide window manner like \cite{zhang2018learningTermTracking, yan2019skimming, xuan2021siamLT}, however, such an approach may need high time cost. For example, the local-global tracker proposed in \cite{xuan2021siamLT, zhang2018learningTermTracking} only achieves 3.8 and 2.7 FPS, respectively, which are almost impossible to use in practical scenarios. Compared with these works, our proposed tracker utilizes a joint local and global search strategy based on a carefully designed target-aware attention network which can bring more robust and accurate tracking results.

\textbf{Visual Attention for Tracking: }
To handle the influence of video noises and/or tracker noises in extremely challenging conditions, there are several attempts to combine attention maps with visual tracking. 
Choi \emph{et al.} \cite{Choi2016Visual} present an attention-modulated visual tracking algorithm that decomposes an object into multiple cognitive units and trains multiple elementary trackers to modulate the distribution of attention based on various features and kernel types. 
Han \emph{et al.} \cite{Hong2015Online} propose an online visual tracking algorithm by learning a discriminative saliency map using CNN. They also directly search the target object from attention locations. 
He \emph{et al.} \cite{he2016KPSRT} propose a robust tracker based on key patch sparse representation (KSPR) to address the issues of partial occlusion and background information. Wang \emph{et al.} \cite{wang2021deepmta} also adopt the dynamic attention network for multi-trajectory analysis to achieve robust tracking. Choi \emph{et al.} \cite{choi2017attentional} develop a deep attentional network to select subset of filters for robust and efficient tracking. Chu \emph{et al.} propose a spatial-temporal attention mechanism (STAM) to handle the drift caused by occlusion and interaction among targets in \cite{chu2017online}. Yang \emph{et al.} \cite{yang2018dmn} also introduce the attention model into their memory network for tracking. Li \emph{et al.} \cite{li2020robustTrack} introduce the channel attention and focal loss to enhance feature representation learning.    
These approaches emphasize attentive features and resort to additional attention modules to generate feature weights. However, the feature weights learned in a single frame are unlikely to enable classifiers to concentrate on robust features over a long temporal span. Instead, our proposed target-aware attention network take continuous video frames and initial target object as inputs and predict frame-specific attention maps for candidate search location mining. Based on the joint local and global search framework, our tracker can attain good performance on both short-term and long-term tracking benchmark datasets.

\textbf{Generative Adversarial Learning: }
Generative Adversarial Networks (GAN) \cite{Goodfellow2014Generative} is a system of two neural networks contesting with each other in a zero-sum game framework. Although many works have been proposed in the object detection area, however, seldom of researchers consider using GAN to handle the issues in visual tracking task \cite{zhao2018adversarial, SongYiBing_2018_CVPR, zhao2020antidecay}. 
Zhao \emph{et al.} \cite{zhao2020antidecay} propose the antidecay LSTM for Siamese tracking which is trained under adversarial learning framework. Song \emph{et al.} \cite{SongYiBing_2018_CVPR} conduct adversarial learning by dropping partial feature maps to help improve the robustness of their tracker. Different from these works, we propose to utilize GAN conditioned on input video frames and initial target objects for target-aware attention estimation for global search.

\section{The Proposed Approach} 

In this section, we will first give an overview of our proposed tracker in section \ref{Overview}. Then, we dive into the details of our target-aware attention network, including the generator and two discriminators. After that, we will talk about the objective functions of our neural network and the details in the tracking phase.

\subsection{Overview} \label{Overview}
As shown in Fig. \ref{trackerpipeline}, our proposed tracker contains two main modules: \emph{the baseline tracker} and \emph{target-aware attention network (TANet)}. In more detail, our TANet is developed based on the framework of generative adversarial network which contains one generator and two discriminators, as shown in Fig. \ref{pipeline}. Our generator takes the continuous video frames and target templates as input and extracts their features with 3D CNN and 2D CNN respectively. Then, the features are concatenated along the channel dimension and sequentially transformed into corresponding attention maps with an upsample network. With the generated sequential attention maps, we utilize combined loss functions of BCE loss, appearance adversarial loss, and motion adversarial loss for the optimization of the generator. In the testing phase, only the generator is used for global attention prediction. Given the global attention maps, we integrate it with baseline tracker for joint local and global search for robust object tracking. More details about these modules can be found in the following subsections.

\subsection{Network Architecture}   

Generally speaking, our TANet contains three modules: the attention generator, the appearance discriminator, and the motion discriminator. We give a detailed introduction to these modules in the following subsections respectively.

\subsubsection{The Generator}
Given the continuous video frames $\{I_{i-1}, I_i, I_{i+1}\}$ ($i$ is the index of video frames, we take three continuous frames as an example to describe our algorithm) and target object initialized in the first frame, we first extract their features using 3D convolutional neural network \cite{tran2018closer3DCNN} and ResNet-18 \cite{he2016identityResNet} respectively. Specifically, we resize the video frames into $300 \times 300$, then feed them into two networks to obtain its motion feature $F_{m}$ and appearance feature $F^i_{a}$. For the target template, we can obtain its feature map $F_{t}$ by feeding the resized image into the ResNet-18. Then, we concatenate the feature maps along the channel dimension, formally, we have: 
\begin{equation}
\label{encodingFeature}
F^i_{encode} = [F_{m}, F^i_{a}, F_{t}]; 
\end{equation}
where $[ , ]$ denotes concatenate operation along the channel dimension. With aforementioned operations, we have feature maps of corresponding input $\{F^{i-1}_{encode}, F^{i}_{encode}, F^{i+1}_{encode}\}$. Then, we use an upsample network to increase the resolution of generated attention maps recurrently. In another word, we use the upsample network to transform the convolutional features into attention maps along the temporal direction, as shown in Fig. \ref{pipeline}. In more detail, the upsample network is composed of five groups of upsample module, and each upsample module consists of three transposed convolutional layers and one upsample layer. It is worthy to note that the upsample network is shared in this procedure.

Given the generated target-aware attention maps, we can directly utilize the \emph{pixel-wise} loss function (BCE used in this paper) to measure its distance with the ground truth mask. However, as noted in \cite{luo2018macroAdversarial, wang2019GANTrack, Pan2017SalGAN}, BCE loss merely penalizes the false prediction on every pixel without explicitly modeling the correlation among adjacent pixels. Besides, it also unable to keep the consistency between continuous video frames. Therefore, in this paper, we resort to the appearance and motion discriminators to joint configure of label variables and enforce higher-level spatial and temporal consistency. The details about the used discriminators can be found in the following subsections respectively.

\subsubsection{The Appearance Discriminator}
As mentioned in our conference paper \cite{wang2019GANTrack}, the appearance discriminator can be used to help guide the attention generation from the perspective of global spatial view. Many works are developed based on this idea, including shadow detection \cite{nguyen2017shadowDetectionGAN, JifengWang_2018_CVPR}, saliency detection \cite{wang2019qualityRGBTsaliency, Pan2017SalGAN} and segmentation \cite{luc2016semanticGAN}. As shown in Fig. \ref{pipeline}, the discriminator consists of convolutional layers and fully connected layers like standard CNN for image classification. The detailed network architecture can be found in Table \ref{appDIS}. 

\begin{table}[t]
\center
\caption{The detailed architecture of appearance discriminator.}\label{appDIS}
\begin{tabular}{l|cccccc}
\hline \toprule [1 pt]
Layer   				&Kernel  					&Stride	&Activation	&\# Input  &\# Out\\
\hline
2D Conv				&7$\times$7		 		&2				&ReLU			&4   			&96\\
Pooling 			&2$\times$2				&2				&-      			&-			&-\\
2D Conv				&3$\times$3				&2				&ReLU			&96			&256\\
Pooling 			&2$\times$2				&2				&-      			&-			&-\\
2D Conv				&3$\times$3				&1				&ReLU			&256		&256\\
Fc1	  				&-							&-				&ReLU			&57600      &256\\
Fc2	  				&-							&-				&ReLU			&256	    &1\\
\hline \toprule [1 pt]
\end{tabular}
\end{table}

\subsubsection{The Motion Discriminator} \label{motionDis} 
The joint utilization of BCE and appearance discriminator can enable us to attain a good single feature map. However, we find that the attention predicted in continuous video frames may be varying significantly, as illustrated in Fig. \ref{motivation}. As we all know, there is only one true target object in each frame for single object tracking and the variation between continuous frames is relatively small. With this in mind, we introduce an external motion discriminator to ensure consistency in the temporal view. As shown in Fig. \ref{pipeline}, we concatenate the predicted several attention maps with their corresponding images and feed them into the motion discriminator for the temporal adversarial learning.

For the detailed network architecture, we first use a set of 3D convolutional layers to encode the input video clips.  Then, we use one 2D convolutional layer and 2 fully connected layers to transform the encoded features into the predicted binary category. We adopt ReLU as the activation layer in the whole motion discriminator. More detailed parameters can be found in Table \ref{motDIS}.

\begin{table}[t]
\center
\scriptsize 
\caption{The detailed architecture of Motion discriminator.}\label{motDIS}
\begin{tabular}{l|cccccc}
\hline \toprule [1 pt]
Layer   				&Kernel  										&Stride			&Padding				&Activation		&\# Input  	&\# Out\\
\hline
3D Conv				&3$\times$3$\times$3		 		&(1, 1, 1)			&(1, 1, 1)				&ReLU				&3   				&64		\\
Pooling 				&1$\times$2$\times$2				&(1, 2, 2)			&-      					&-					&-				&-		\\
3D Conv				&3$\times$3$\times$3				&(1, 1, 1)			&(1, 1, 1)				&ReLU				&64				&128	\\
Pooling 				&2$\times$2$\times$2				&(2, 2, 2)			&-      					&-					&-				&-		\\
3D Conv				&3$\times$3$\times$3				&(1, 1, 1)			&(1, 1, 1)				&ReLU				&128			&256	\\
3D Conv				&3$\times$3$\times$3				&(1, 1, 1)			&(1, 1, 1)				&ReLU				&256			&256	\\
Pooling 				&2$\times$2$\times$2				&(2, 2, 2)			&-      					&-					&-				&-		\\
3D Conv				&3$\times$3$\times$3				&(1, 1, 1)			&(1, 1, 1)				&ReLU				&256			&512	\\
3D Conv				&3$\times$3$\times$3				&(1, 1, 1)			&(1, 1, 1)				&ReLU				&512			&512	\\
Pooling 				&1$\times$2$\times$2				&(2, 2, 2)			&(1, 1, 1)				&-      				&-				&-		\\
3D Conv				&3$\times$3$\times$3				&(1, 1, 1)			&(1, 1, 1)				&ReLU				&512			&512	\\
3D Conv				&3$\times$3$\times$3				&(1, 1, 1)			&(1, 1, 1)				&ReLU				&512			&512	\\
\hline
2D Conv				&3$\times$3								&2					&1						&ReLU				&512			&100\\
Pooling 				&3$\times$3								&2					&-						&-      				&-				&-\\
\hline
Fc1	  					&-											&-					&-						&ReLU				&900      		&256\\
Fc2	  					&-											&-					&-						&ReLU				&256	    	&1\\
\hline \toprule [1 pt]
\end{tabular}
\end{table}

\subsection{Objective Functions}  

Given input video frames $\{I_{i-1}, I_i, I_{i+1}\}$ \footnote{We use $I$ to denote the tube $\{I_{i-1}, I_i, I_{i+1}\}$ for simplicity in later sections.}, target object template $T$, target label map $y$ of shape $1 \times H \times W$ where $W, H$ are width and height of maps, the binary cross-entropy loss function can be formulated as: 
\begin{equation}
\label{bceLoss}
L_{bce} (G) = \sum_{i=1}^{H \times W} -y_{i} log y'_i 
\end{equation}
where $y'_i$ denotes the predicted probability on the $i$-th pixel. $y_{i}$ is the ground truth probability of target object on the $i$-th pixel. More detail, if the $i$-th pixel belongs to target object, $y_i$ = 1, otherwise $y_i$ = 0. 
In our conference paper \cite{wang2019GANTrack}, we combine the pixel-wise loss function with adversarial loss (i.e., the appearance discriminator) to enforce the spatial consistency. The loss function of appearance adversarial loss $L_{GAN}^{a} (G, D_{a})$ can be formulated as: 
\begin{equation}
\label{appGANLoss}
\begin{aligned}
L_{GAN}^{a} (G, D_{a}) =
  & \mathbb{E}_{I, y_i}[log D_{a}(I, y_i)] + \\
  & \mathbb{E}_{I} [log (1-D_{a}(I, G(I, T)))]
\end{aligned}
\end{equation}

To ensure the consistency of predicted continuous attention maps, we extend aforementioned adversarial learning framework by introducing another network, i.e., the motion discriminator $L_{GAN}^{m} (G, D_{m})$. We denote the continuous video frames and corresponding attention maps $[y_{i-1}, y_{i}, y_{i+1}]$  as $I$ and $\hat{y}$ respectively. The motion adversarial loss can be written as: 
\begin{equation}
\label{motionGANLoss}
\scriptsize 
\begin{aligned} 
L_{GAN}^{m} (G, D_{m}) = 
  & \mathbb{E}_{I, \hat{y}}[log D_{m}(I, \hat{y})] + \\ 
  & \mathbb{E}_{I} [log (1-D_{m}(I, [G(I_{i-1}, T), G(I_{i}, T), G(I_{i+1}, T)]))] 
\end{aligned}
\end{equation}
Therefore, we have the final objective function which can be formulated as: 
\begin{equation}
\label{ObjectiveFunc}
L = L_{bce} + \lambda_1 * L_{GAN}^{a} (G, D_{a}) + \lambda_2 * L_{GAN}^{m} (G, D_{m})
\end{equation}
where $\lambda_1$ and $\lambda_2$ are all trade-off parameters. Detailed analysis on the two parameters are given in our experiments.
This function can also be rewritten as: 
\begin{equation}
\label{ObjectiveFuncGAN}
G^*, D_{a}^*, D_{m}^* = \mathop{\arg}\min_{G} \max_{D_{a}, D_{m}}  L(G, D_{a}, D_{m})
\end{equation}
The Eq. \ref{ObjectiveFuncGAN} can be solved by alternate between optimizing $G, D_{a}$ and $D_{m}$ until $L(G, D_{a}, D_{m})$ converges. The training process of our proposed dual-discriminator based generative adversarial learning can be found in Algorithm \ref{algorithm}.

\renewcommand{\algorithmicrequire}{\textbf{Input:}}
\renewcommand{\algorithmicensure}{\textbf{Output:}}
\begin{algorithm}
	\small 
	\caption{Dual-Discriminator based Adversarial Learning for Target-aware Attention Prediction.} 
	\label{algorithm}
	\begin{algorithmic}[1]
		\REQUIRE  Video frame $\{f_1, ... , f_N\}$, Ground truth $\{p_1, ... , p_N\}$, iter = 0, $n_1$=5, $n_2$=3, $\lambda_1=0.2, \lambda_2=0.1$. \\ 
		
		\ENSURE The attention generator $G$. \\ 
		
		\textbf{While} not converged \textbf{do}:  \\ 
		~~\textbf{If}  iter \% $n_1$ == 1: \\  
		~~~~Get M data samples $(I, y)$ \\ 
		~~~~\textbf{Update the Appearance Discriminator $D_{a}$: } \\ 
		~~~~~~~~  $\theta_{appD} = \theta_{appD} - \rho_{a} \sum_{i=1}^{M} \frac{\alpha L_{GAN}^{a}(I_i, y_i)}{\alpha \theta_{appD}}$ \\ 
		~~\textbf{Else If} iter \% $n_2$ == 1: \\   
		~~~~Get M new data samples $(I, y)$ \\ 
		~~~~\textbf{Update the Motion Discriminator $D_{m}$: } \\ 	
		~~~~~~~~ $\theta_{motD} = \theta_{motD} - \rho_{m} \sum_{i=1}^{M} \frac{\alpha L_{GAN}^{m}(I, y)}{\alpha \theta_{motD}}$ \\ 
		~~~~~~~~\textbf{Else}:  \\   
		~~~~Get M new data samples $(I, y)$ \\ 
		~~~~\textbf{Update the Generator $G$: } \\  
		~~~~~~~~  $\theta_{G} = \theta_{G} - \rho_{G} \sum_{i=1}^{M} (\frac{\alpha L_{bce}(I_i, y_i)}{\alpha \theta_{G}} + \lambda_1 * \frac{\alpha L_{GAN}^{a}(I_i, y_i)}{\alpha \theta_{G}} +  \lambda_2 * \frac{\alpha L_{GAN}^{m}(I, y)}{\alpha \theta_{G}})$ \\ 
		~~~~~~~~ \textbf{End If} \\ 
		~~ \textbf{End If} \\ 
		~~iter = iter + 1 \\ 
		\textbf{End While}
	\end{algorithmic}
\end{algorithm}

\begin{figure}[!htb]
\center
\includegraphics[width=3.3in]{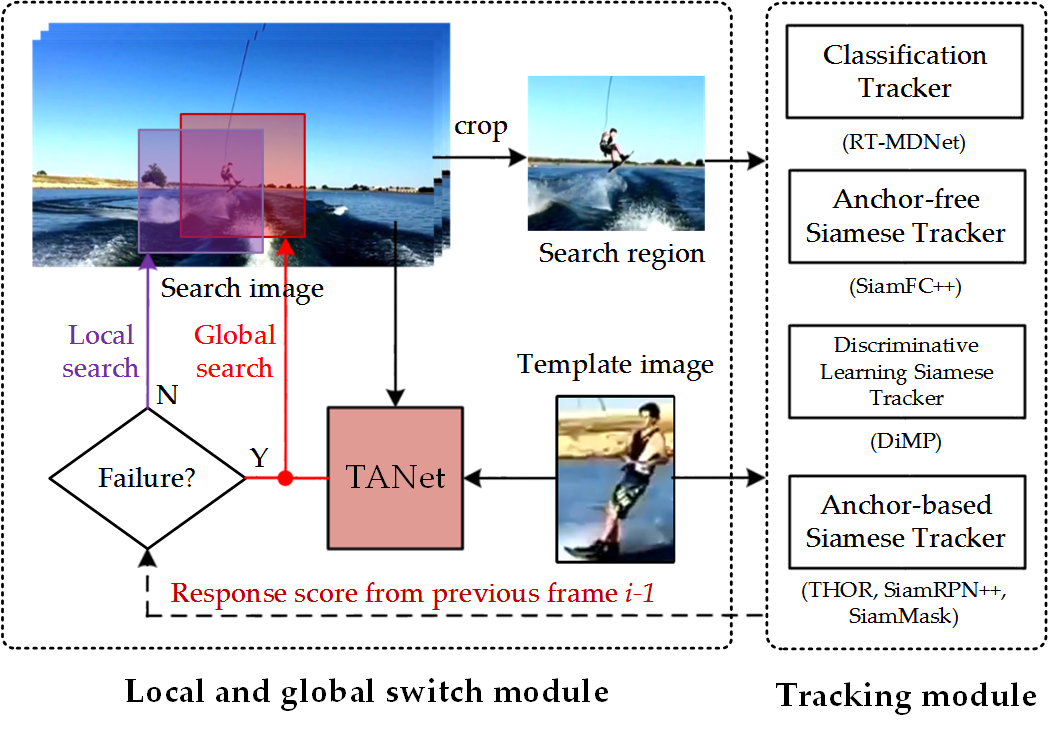}
\caption{Illustration of joint local and global search based visual tracking. }
\label{trackerpipeline}
\end{figure}

\subsection{Joint Local and Global Search for Tracking} 
After we obtain the target-aware attention generator, we can integrate it with recent popular trackers due to our TANet is a plug and play module. In this work, the SiamRPN++ \cite{li2019siamrpn++}, SiamFC++ \cite{xu2020siamfc++}, THOR \cite{THOR2019}, RT-MDNet \cite{jung2018real}, SiamMask \cite{wang2019siammask}, and DiMP \cite{Bhat_dimp} are selected to validate the generalization of our model due to their good performance and high efficiency. In this section, we give a brief introduction to the SiamRPN++ to demonstrate how we conduct joint local and global search for visual tracking based on this tracker. The integration of our module with other trackers shares similar ideas. Specifically, the SiamRPN++ can locate the target object in the search region from the current frame, given the target template in the first frame. The two images are fed into a shared backbone network and the obtained features are processed by some non-shared neck layers and fused by depth-wise correlation operation. Then, the RPN head layers is used to predict the classification map and regression map. The regression branch involves four regression factors, including the center offset and scale changes. The position with the highest classification score is treated as the optimal target location of the current frame. For the detailed introduction of SiamRPN++ and other used trackers, please refer to their original papers.

\begin{figure*}[!htb]
\center
\includegraphics[width=7in]{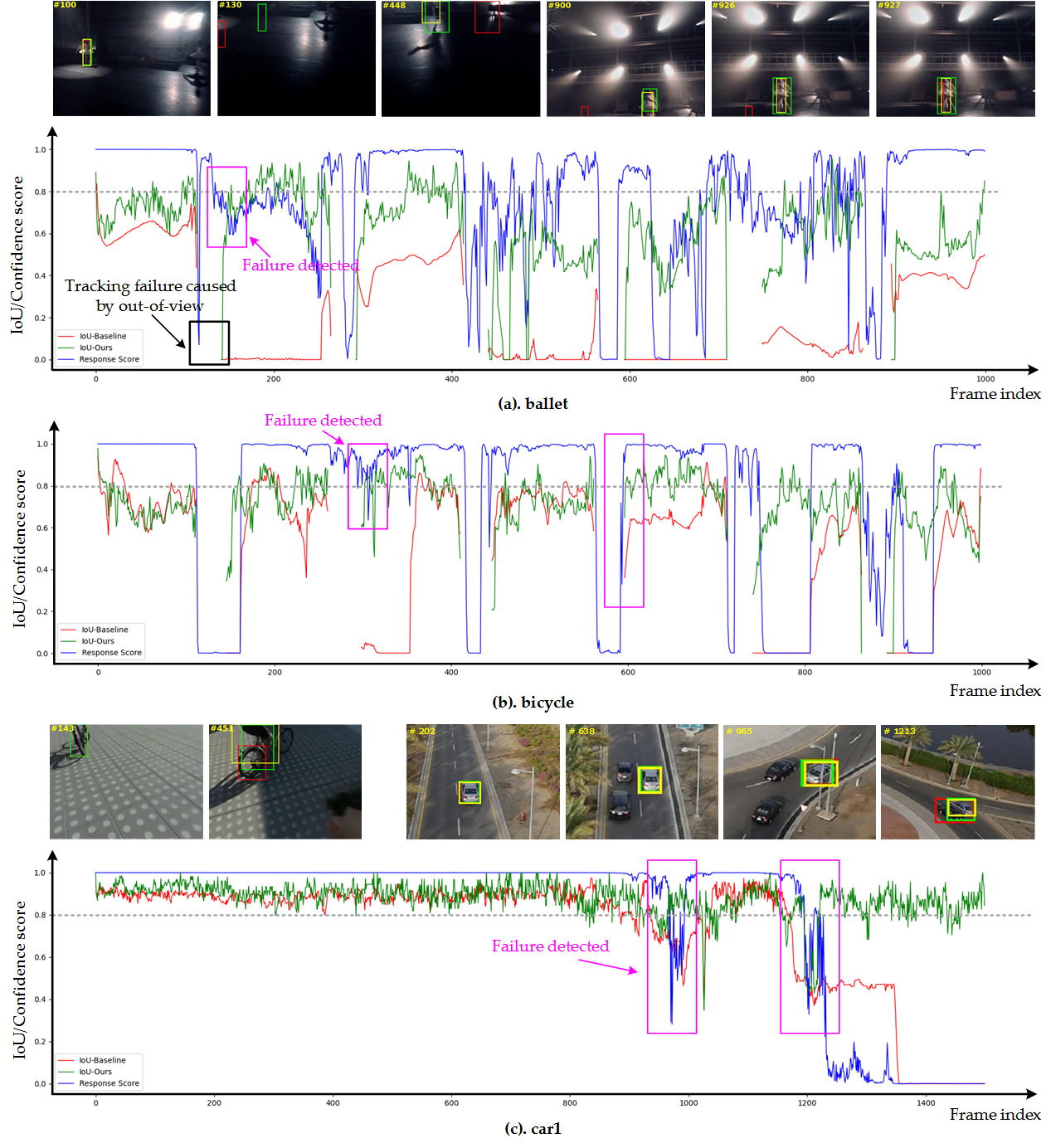}
\caption{Illustration on relations between IoU score and response score from tracker (DiMP is used as baseline tracker for the visualization on selected video \emph{ballet} from VOT2018LT dataset). The discontinuity between the lines is caused by the out-of-view of the target object. The red, green, and yellow BBox visualized in the video frames are tracking results of DiMP, Ours, and ground truth, respectively.}  
\label{iou_plot}
\end{figure*}

Another part of our tracking system is the local and global search module, which plays an important role when facing challenging factors like fast motion, out-of-view, heavy occlusion, etc. Given the tracking results of the previous frame, we adopt the SiamRPN++ tracker to conduct local search if the response score is high (larger than a pre-defined threshold). When the challenging factors occurred, the response score will be unstable (lower than the threshold in continuous video frames) and the tracker may fail to locate the target object accurately. Therefore, we can use the global search region provided by our TANet at this moment, as shown in Fig. \ref{trackerpipeline}. For the next frame, we still adopt local search for tracking and also will switch into global search if failure is detected. Similar operations are executed until the end of the current video sequence. We give a visualization on the IoU score and response score to help readers better understand our motivation and detection of tracking failure. As shown in Fig. \ref{iou_plot}, the red, blue, and green lines are used to denote the IoU value of baseline tracker DiMP \cite{Bhat_dimp}, IoU value of ours, and confidence (response) score of DiMP tracker. We can find that the confidence score can reflect the quality of tracking procedure to some extent which is intuitive and easy to implement. Therefore, we adopt the confidence score as the criterion to switch between local and global tracking. We provide a demo video for the comparison between the baseline tracker and our tracking results which can be found at: \url{https://youtu.be/F2rrgLVS4nE}. 
This procedure can also be found in Algorithm \ref{trackAlgo}.

\renewcommand{\algorithmicrequire}{\textbf{Input:}}
\renewcommand{\algorithmicensure}{\textbf{Output:}}
\begin{algorithm}
	\small 
	\caption{Joint Local and Global Search based Framework for Robust Single Object Tracking} 
	\label{trackAlgo}
	\begin{algorithmic}[1]
		\REQUIRE  Test video sequence $\{f_1, ... , f_N\}$ and initial BBox, Baseline Tracker, and TANet, $\beta_1$ = 0.8, $\beta_2$ = 5, fail count $\beta$ = 0. \\ 
		
		\ENSURE Tracking Results. \\ 
		
		\textbf{For} each frame $f_i$ in $\{f_2, ... , f_N\}$: \\ 
		~~ \textbf{If} response score of frame $f_{i-1}$ larger than $\beta_1$: \\  
		~~~~ Conduct $local$ search with Baseline Tracker; \\ 
		~~ \textbf{Else If} $\beta$ < $\beta_2$: \\ 
		~~~~ Conduct $local$ search with Baseline Tracker; \\ 
		~~~~ $\beta$ = $\beta$ + 1; \\ 
		~~~~~~~~ \textbf{Else}: \\ 
		~~~~ Predict $global$ search region with TANet; \\
		~~~~ Run Baseline Tracker according to $global$ search region; \\ 
		~~~~ $\beta$ == 0; \\
		~~~~~~~~ \textbf{End If} \\ 
		~~ \textbf{End If} \\ 
		\textbf{End For}
	\end{algorithmic}
\end{algorithm}

\section{Experiment}

\subsection{Datasets and Evaluation Metrics} 
For the training of our TANet, we randomly select 500 videos from train subset of GOT-10k dataset. We test our model on both short-term and long-term tracking datasets, including: OTB-100 \cite{wu2015object}, GOT-10k \cite{huang2019got}, TrackingNet \cite{muller2018trackingnet}, VOT2018LT \cite{vot2018}, VOT2019LT \cite{kristanvot2019} and LaSOT \cite{fan2019lasot}.

\textbf{Success Rate (SR) and \textbf{Precision Rate (PR)}} are adopted for the evaluation of LaSOT, OTB-100 and GOT-10k dataset. Specifically, the SR is used to measure interaction-over-union overlap between the predicted bounding box and ground truth. The frame will be treated as success if the IoU is larger than a certain threshold. The PR focus on the center of the predicted bounding box and the ground truth bounding box. The tracking will be considered as success if the distance is under some threshold. The \textbf{Average Overlaps (AO)} is also adopted for the evaluation of GOT-10k which denotes the average of overlaps between all ground truth and estimated bounding boxes. As noted in \cite{huang2019got}, the AO is recently proved to be equivalent to the area under curve (AUC) metric employed in OTB-100 \cite{wu2015object}, TrackingNet \cite{muller2018trackingnet}, and LaSOT \cite{fan2019lasot}. For the VOT2018LT and VOT2019LT datasets, we adopt the \textbf{Precision}, \textbf{Recall}, and \textbf{F1-score} as the evaluation criteria. Specifically, the definition of these metrics are: 
\begin{equation}
\label{Precision}
\small  
Precision = \frac{TP}{TP + FP}, ~~~~Recall = \frac{TP}{TP + FN}
\end{equation}
\begin{equation}
\label{Precision}
\small  
F1-score =  \frac{2 \times Precision \times Recall}{Precision + Recall}
\end{equation}
where TP, FP, and FN are used to denote the True Positive, False Positive and False Negative, respectively.

\subsection{Implementation Details}
In the training phase, the ground truth mask used for the training of our TANet can be directly obtained from existing tracking datasets. Following our conference paper \cite{wang2019GANTrack}, we fill the foreground and background regions with white pixels and black pixels respectively. It is also worthy to note that we recurrently predict continuous attention maps for each video frame in the training phase. In the testing stage, we directly predict one attention map for each frame to decrease the cost. Moreover, we monitor the failure of tracking procedure based on previous response scores $\beta_1$. We will switch to global search if the response score is lower than the given threshold for continuous $\beta_2$ frames. The related analysis on the two parameters can be found in later ablation studies.

Adagrad \cite{duchi2011adaptive} is adopted for the optimization of our TANet. The learning rate for the generator and two discriminators are all $1e-4$. The experiments are conducted on a server with 8 $\times$ TITAN Xp. For the weights used in our baseline tracker, we directly use the pre-trained model released by the authors. Our code is implemented based on PyTorch \cite{paszke2019pytorch} and will be released at \url{https://github.com/wangxiao5791509/LGSearch_DDGAN_PyTorch}.

\subsection{Comparison on Public Benchmarks} 

Six popular tracking benchmark datasets are used for the performance evaluation, including three short-term and three long-term tracking datasets. 

\textbf{Results on OTB100 \cite{wu2015object}: }   
OTB100 is a very popular benchmark in the visual tracking community which is consisted of 100 video sequences and proposed in 2015, also named OTB-2015. We can find that the baseline tracker THOR achieves 0.787/0.641 on the PR/SR on this benchmark according to Table \ref{OTBGOTTrackNetresults}. With the help of our attention maps, we can conduct a global search and attain better results. Specifically, we achieve 0.791/0.646 on this benchmark and these results are significantly better than the baseline approach. This experimental result validated the effectiveness of our model.

\begin{table*}[htp!]
\center
\scriptsize  
\caption{Results on OTB100, GOT-10k, TrackingNet, VOT2018-LT and VOT2019-LT dataset. The best results are highlighted in \textbf{bold}.} \label{OTBGOTTrackNetresults}
\begin{tabular}{c|cccccc|cccccccc}
\hline \toprule [1 pt]
\textbf{OTB100}	&{SiamFC \cite{bertinetto2016fully}}     &{ PGNet \cite{liao2020pgnet}}    &Ocean \cite{zhang2020ocean}     &{PTAV \cite{fan2017ptav}}       &SiamRCNN \cite{voigtlaender2020siamRCNN}          &{CREST \cite{song2017crest}}       &{THOR \cite{THOR2019}}        &{Ours}  \\
PR/SR               &0.771/0.582  &0.892/0.691   &\textbf{0.920}/0.684  &0.848/0.634   &0.891/\textbf{0.701}    &0.838/0.623   &0.787/0.641   &0.791/0.646 		\\  
\hline \toprule [1 pt]
\textbf{GOT-10k} 	&ROAM \cite{yang2020roam}	&SiamRCNN \cite{voigtlaender2020siamRCNN}   	& D3S \cite{lukezic2020d3s} &SiamFC \cite{bertinetto2016fully}	&ATOM \cite{danelljan2019atom}	&RT-MDNet \cite{jung2018real}  &SiamFC++ \cite{xu2020siamfc++}	&Ours  \\ 
$AO/SR_{0.50}$     &0.465/0.532		&\textbf{0.649}/0.728 	&0.597/0.676 	&0.348/0.353   &0.556/0.634 &0.342/0.356 &0.595/0.695 &0.607/\textbf{0.737}		\\ 
\hline \toprule [1 pt]
\textbf{TrackingNet}    &SiamFC \cite{bertinetto2016fully}		&MDNet \cite{Nam2015Learning} 	&ATOM \cite{danelljan2019atom} 	&SiamRPN++ \cite{li2019siamrpn++}   &DiMP \cite{Bhat_dimp}	&DaSiam-RPN \cite{zhu2018Dist-RPN} &SiamFC++ \cite{xu2020siamfc++}	&Ours  \\ 
Precision    			&0.533		&0.565 	&0.648 	&0.694      &0.687 		&0.591 		    &0.646 	&\textbf{0.711}		\\ 
Norm. Prec.     	    &0.666		&0.705 	&0.771 	&0.800      &0.801 		&0.733 		    &0.758 	&\textbf{0.813}		\\ 
Success (AUC)    	&0.571		&0.606 	&0.703 	&0.733      &0.740 		&0.638 		    &0.712 	&\textbf{0.746}		\\ 
\hline \toprule [1 pt]
\textbf{VOT2018-LT} 	&{PTAV+ \cite{fan2017ptav}}  &{SYT}  &{LTSINT\cite{Tao2016Siamese}}  &{MMLT} \cite{lee2018MMLT}  &{DaSiam-LT} \cite{zhu2018Dist-RPN}      &{MBMD} \cite{zhang2018learningTermTracking}   &SiamRPN++ \cite{li2019siamrpn++}  &{Ours} \\
Precision 		&0.595     &0.520      &0.566      &0.574    &0.627      &0.634       &0.646       &\textbf{0.649}        \\
Recall 			&0.404     &0.499      &0.510      &0.521    &\textbf{0.588}      &\textbf{0.588}       &0.419       &0.535        \\
F1-score 		&0.481     &0.509      &0.536      &0.546    &0.607      &\textbf{0.610}       &0.508       &0.586       \\
\hline \toprule [1 pt]
\textbf{VOT2019-LT} 	 &{ASINT }  &{CooSiam}  &Siamfcos-LT  &SiamRPNs-LT \cite{li2018siamRPN}     &mbdet   &SiamDW-LT \cite{zhang2019siamDW}  &{SiamRPN++ \cite{li2019siamrpn++}}  &{Ours} \\
Precision 		 &0.520      &0.566      &0.574    &0.627      &0.634       &\textbf{0.649}      &0.627   &0.628         	  \\
Recall 			 &0.499      &0.510      &0.521    &0.588      &0.588       &\textbf{0.609}      &0.399   &0.437             \\
F1-score 		&0.509      &0.536      &0.546    &0.607      &0.610        &\textbf{0.629}      &0.488   &0.515             \\
\hline \toprule [1 pt]
\end{tabular}
\end{table*}

\textbf{Results on GOT-10k \cite{huang2019got}:} 
GOT-10k is a recently released large-scale tracking benchmark that contains 180 challenging videos for testing. As shown in Table \ref{OTBGOTTrackNetresults}, the baseline method SiamFC++ (backbone: GoogLeNet) achieves 0.595/0.695 on the AO/SR respectively, while our proposed TANet can improve them to 0.607/0.737. Our tracking results are better than most of the compared algorithms including ROAM \cite{yang2020roam}, D3S \cite{lukezic2020d3s}, and ATOM \cite{danelljan2019atom}. It is worthy to note that the SiamRCNN \cite{voigtlaender2020siamRCNN} is also a joint local and global search based tracker which is developed based on re-detection scheme. We achieve better results on SR when compared with this tracker. In addition, we also integrate our TANet into DiMP \cite{Bhat_dimp} in later parameter analysis, specifically, the baseline DiMP (SuperDiMP version) achieves 0.666/0.777 on $AO/SR_{0.50}$ and we can further improve it to 0.684/0.799. This result is significantly better than the compared trackers including SiamRCNN  \cite{voigtlaender2020siamRCNN}, D3S \cite{lukezic2020d3s}, etc. Therefore, extensive experimental results on this benchmark  fully demonstrate the advantage and effectiveness of our model.

\textbf{Results on TrackingNet \cite{muller2018trackingnet}:} 
TrackingNet contains 511 video sequences for testing, as shown in Table \ref{OTBGOTTrackNetresults}, we can find that the baseline tracker SiamFC++ (backbone: AlexNet) achieves 0.646/0.758/0.712 on the PR/Normalized-PR/SR respectively which are already better than some recent strong trackers like SiamRPN++ and ATOM. We can further improve it to 0.711/0.813/0.746 with our TANet which also demonstrates the effectiveness of our model and the importance of global search in visual tracking. Our results are also better than DiMP \cite{Bhat_dimp} which is a strong tracker proposed in 2019. Due to the baseline tracker already attained good results, and the video sequences in TrackingNet are nearly all short-term videos, therefore, our improvement is relatively smaller than on long-term datasets.

\textbf{Results on VOT2018-LT \cite{vot2018}:}
VOT2018LT contains 35 challenging video sequences with a total length of 146,817 frames. As noted in \cite{yan2019skimming}, each video contains on average 12 long-term target disappearances, each lasting on average 40 frames. This dataset employs the precision/recall/F1-score as the evaluation criteria. As reported in Table \ref{OTBGOTTrackNetresults}, we can find that the baseline tracker SiamRPN++ achieves 0.646/0.419/0.508 on these three metrics, while we obtain 0.649/0.535/0.586. This experiment fully demonstrates the effectiveness of our model for long-term visual tracking task.

\begin{figure*}[!htb]
\center
\includegraphics[width=7in]{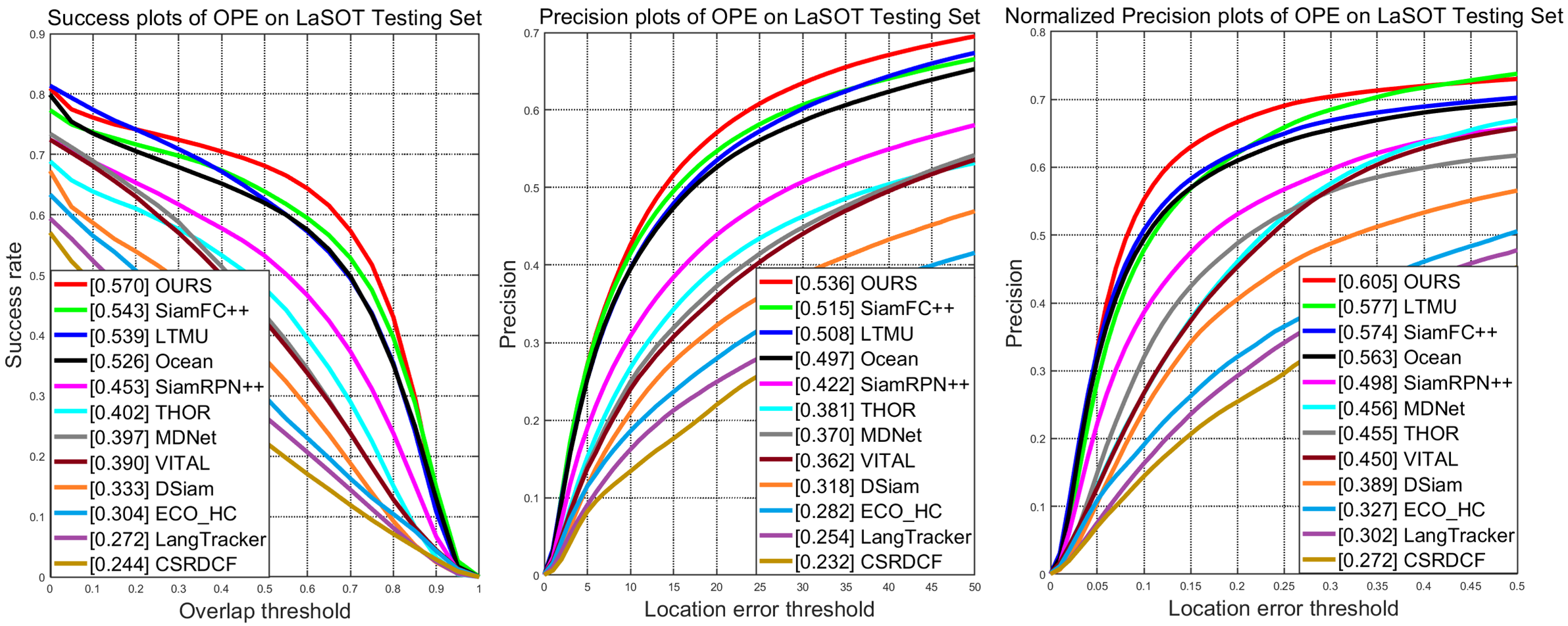}
\caption{Tracking results on the LaSOT dataset.}  
\label{ganTANet_lasot}
\end{figure*}

\textbf{Results on VOT2019-LT \cite{kristanvot2019}}
The VOT2019LT dataset consists of 50 long-term video sequences (215,294 frames). It introduces 15 more difficult videos with uncommon targets, such as \emph{boat}, \emph{bull}, and \emph{parachute}, etc, based on VOT2018LT. It adopts the same evaluation protocol with VOT2018LT. As shown in Table \ref{OTBGOTTrackNetresults}, we can find that the baseline tracker SiamRPN++ achieves 0.627/0.399/0.488 on this benchmark, and we can improve it to 0.628/0.437/0.515 on the precision/recall/F1-score respectively.

\textbf{Results on LaSOT \cite{fan2019lasot}:} 
LaSOT is the largest dataset for long-term tracking which contains 1400 videos (Protocol-I and protocol-II employ all the videos and 280 videos for testing respectively). In this paper, we adopt the protocol-II for the evaluation of our tracker and compared trackers including CSRDCF \cite{lukezic2017discriminative}, Lang-Tracker \cite{wang2018describe}, ECO \cite{Danelljan2016ECO}, DSiam \cite{guo2017DSiam}, VITAL \cite{SongYiBing_2018_CVPR}, THOR \cite{THOR2019}, MDNet \cite{Nam2015Learning}, SiamRPN++ \cite{li2019siamrpn++}, Ocean \cite{zhang2020ocean}, SiamFC++ \cite{xu2020siamfc++}, LTMU \cite{dai2020high}. As shown in Fig. \ref{ganTANet_lasot}, we can find that the baseline tracker SiamFC++ achieves 0.574/0.543 on the PR/SR. Our proposed TANet can further improve them to 0.605/0.570 on the two evaluation metrics which outperform the baseline tracker by +3.1/2.7 points. This experiment fully validates the effectiveness of the global search provided by our TANet. Compared with other recent state-of-the-art trackers, we can find that our tracker also outperforms their results by a large margin. Specifically, we outperform the LTMU by +2.8/3.1 which is a recent strong long-term tracker proposed in \cite{dai2020high}. Our results are also better than recent strong anchor-free tracker Ocean \cite{zhang2020ocean}, anchor-based tracker SiamRPN++ and THOR, which all demonstrate the advantages of our proposed module. Similar conclusions can also be drawn from the normalized PR results in Fig. \ref{ganTANet_lasot}.

\begin{table}[!htp]
\center
\scriptsize 
\caption{Ablation study of loss functions on VOT2018LT.} 
\label{CMAnalysis}
\begin{tabular}{c|ccccc|c} 
\hline \toprule [1 pt]
&\textbf{+C3D}  &\textbf{+ResNet} &\textbf{+BCE} &\textbf{+AppD}  &\textbf{+MotD}  &\textbf{F1-score} \\
\hline 
Baseline      &  	&     &  	&  &           &0.508    \\
\hline 
Algorithm-1  &\cmark  	&     &\cmark  	&       &               &0.542    \\
\hline 
Algorithm-2  &  	&\cmark     &\cmark  	&        &          &0.565    \\
\hline 
Algorithm-3  &\cmark  	&\cmark     &\cmark  	&        &      &0.572    \\
\hline 
Algorithm-4  &\cmark  	&\cmark     &\cmark  	&\cmark  &      &0.575    \\
\hline 
Algorithm-5  &\cmark  	&\cmark     &\cmark  	&  &\cmark      &0.577    \\
\hline 
Algorithm-6  &\cmark  	&\cmark     &\cmark  	&\cmark  &\cmark   &0.579    \\
\hline 
Algorithm-6s  &\cmark  	&\cmark     &\cmark  	&\cmark  &\cmark   &0.586    \\
\hline \toprule [0.8 pt]
\end{tabular}
\end{table}

\subsection{Ablation Study}

In this section, we will first give a component analysis of our model to check the effectiveness of each component. Efficiency analysis and attribute analysis are also reported in subsequent subsections.

\subsubsection{Component Analysis on Encoding Network}  
In this subsection, we implement three kinds of encoding networks to validate the effectiveness of each module for final tracking, i.e., Algorithm-1, Algorithm-2, and  Algorithm-3 in Table \ref{CMAnalysis}. These three models encode input data with C3D only, ResNet only, and both of them, respectively. As we can see from Table \ref{CMAnalysis}, when we remove the ResNet network or C3D module, the performance all dropped compared with both of them used. This demonstrates that C3D and ResNet modules all contribute to our final tracking results. Besides, we can also find that the tracking results with attention modules are significantly better than the baseline method. This fully validates the effectiveness of our joint local and global search framework based on the proposed TANet.

\subsubsection{Component Analysis on Loss Functions}
As shown in Table \ref{CMAnalysis}, we implement the following component analysis to check the contributions of each component in our loss functions: 
$(\emph{i})$ {+BCE:} the binary cross-entropy loss function is used for the training of TANet; 
$(\emph{ii})$ {+AppD}: the appearance discriminator is used; 
$(\emph{iii})$ {+MotD}:  the motion discriminator is used. 
It is worthy to note that the SiamRPN++ is selected as the baseline approach which uses AlexNet as the backbone network.

\textbf{Effects of Global Search.} 
As we can see from Table \ref{CMAnalysis}, the baseline tracker SiamRPN++ achieves 0.508 of the F1-score on the VOT2018LT dataset. When integrated with our target-aware attention maps, we can improve the baseline significantly by exploring joint local and global search scheme. Specifically, When all the models used, i.e., the Algorithm-6 in Table \ref{CMAnalysis}, our final result achieves 0.586,  which fully demonstrates that our joint local and global search is very helpful for tracking task. More experimental results in section \ref{generalization} (generalization study) also validated this conclusion.

\textbf{Effects of Appearance Adversarial Learning.}  
To obtain better attention maps, we introduce the adversarial learning strategy to model the relations between different pixel values. In our conference paper, we design a simple appearance discriminator and jointly optimize the generator with BCE loss, i.e., the Algorithm-4 in Table \ref{CMAnalysis}. We can find that adversarial training also contributes to our final performance compared with Algorithm-3.

\textbf{Effects of Temporal Adversarial Learning.} 
In addition to the aforementioned appearance adversarial learning, we propose a novel motion discriminator to ensure the consistency of attention maps between nearby frames. As shown in Table \ref{CMAnalysis}, when jointly utilize these two discriminators, we can obtain 0.579 on this benchmark. It is also worthy to note that Algorithm-4 and Algorithm-5 are all better than the non-adversarial version Algorithm-3. We can obtain better results when more training data used, as shown in Table \ref{CMAnalysis}, the Algorithm-6s achieve 0.586 on this benchmark. These experiments validate that our model can obtain better results with the help of adversarial learning.

\begin{table*}[!htp]
\center
\scriptsize 
\caption{Efficiency analysis on OTB2015 dataset.}  
\label{EfficiencyAnalysis}
\begin{tabular}{cc|cc|cc|cc} 
\hline \toprule [0.8 pt]
SiamRPN++ (AlexNet | ResNet50) & Ours &DiMP (ResNet50) &Ours  &RT-MDNet (CNN-3)  &Ours  &SiamFC++ (GoogleNet)  &Ours \\ 
\hline 
259 | 47 FPS  	&134 | 35 FPS    &31 FPS  &29 FPS  &50 FPS  &44 FPS  &103 FPS  &62 FPS   \\
\hline \toprule [0.8 pt]
\end{tabular}
\end{table*}

\subsubsection{Efficiency Analysis} 
The dual-discriminators are only used for adversarial learning in the training phase, that is to say, the two modules will be removed in the tracking phase which does not increase the time cost. In addition, due to the switch between local search and global search, the target-aware attention model is only employed when the confidence score is rather low. Hence, the overall efficiency of baseline tracker can be ensured as much as possible. 
For each video, we resize the input frames into a fixed scale ($300 \times 300$ used in this paper), therefore, the time spent on the attention prediction is fixed. 
Specifically, the TANet needs to spend 0.18 seconds for each frame when tested on a server with CPU I9 and GPU RTX2080TI. When integrating the TANet with SiamRPN++ (AlexNet used as the backbone), our tracker can run at 134 FPS (frames per second) on the whole OTB2015 dataset. If we use ResNet50 as the backbone, we can run at 35 FPS, as shown in Table \ref{EfficiencyAnalysis}. In addition, we can draw similar conclusions when combining our TANet with other baseline trackers, like RT-MDNet and SiamFC++. These experimental results fully demonstrate the efficiency of our tracker.

\subsubsection{Attribute Analysis}
\begin{figure*}[!htb]
\center
\includegraphics[width=7in]{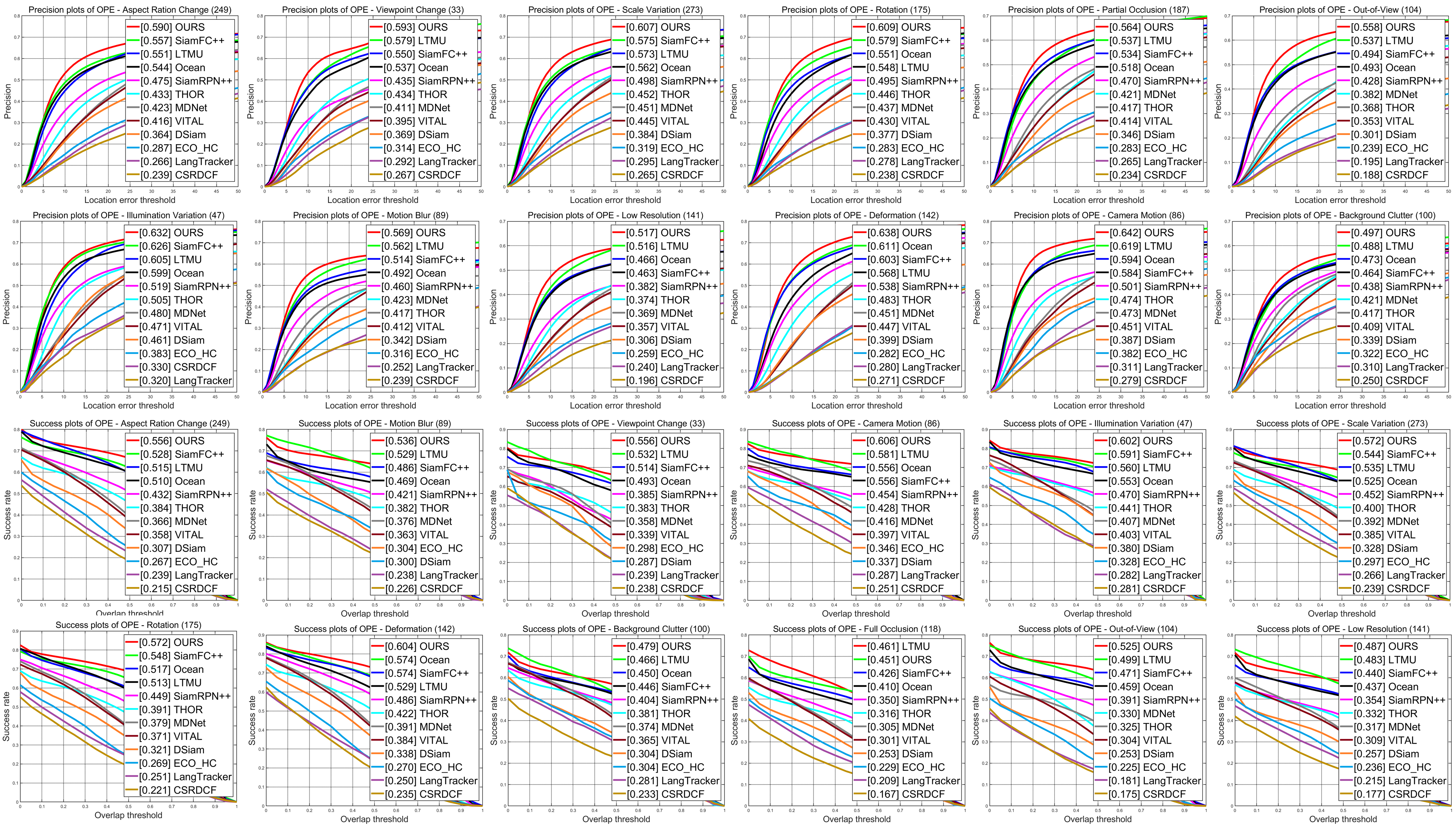}
\caption{Attribute analysis of PR and SR on the LaSOT dataset. Best viewed by zooming in. } 
\label{ganTANet_PRSR_attribute}
\end{figure*}

In this subsection, we report the tracking results under various challenging factors to demonstrate the effectiveness and robustness against these issues. As shown in Fig. \ref{ganTANet_PRSR_attribute}, it is easy to find that our proposed tracker (based on SiamFC++) achieves the best tracking performance on most of the mentioned challenging factors, including \emph{aspect ration change}, \emph{viewpoint change}, \emph{scale variation}, \emph{illumination variation}, \emph{motion blur}, \emph{low resolution}, \emph{deformation}, \emph{rotation}, \emph{camera motion}, \emph{partial occlusion}, \emph{background clutter} and \emph{out-of-view}. For the \emph{full occlusion} and \emph{fast motion}, the LTMU achieves better results than ours. This maybe caused by naive switch policy used in our tracker, i.e., we switch into global search if the tracking score is lower than given threshold for continuous frames. In our future works, we will consider to learn a specialized network for adaptive switch to get better tracking results.

\subsection{Parameter Analysis} 

In our experiments, two parameters are very important for our global search, i.e., the switching threshold $\beta_1$ and failure threshold $\beta_2$. We report the tracking results with different thresholds in this subsection. The baseline tracker used in this experiment is the SiamMask \cite{wang2019siammask}. 

For the switch threshold $\beta_1$, we think the state of tracking procedure is failure if the response score $\beta_2$ is lower than a pre-defined value for continuous $\beta_1$ frames. We fix $\beta_2$ as 0.8, and set the $\beta_1$ as 4, 6, 8, 10, 12. As reported in Table \ref{param_plpt}, we can get 0.450/0.543, 0.453/0.545, 0.455/0.548, 0.455/0.547, 0.454/0.545 on the AO/SR on the GOT-10k dataset. We can find that the overall results are better when the $\beta_1$ is in the range of [8, 10].

For the failure threshold $\beta_2$, it is used to determine whether the current frame is failing or not. In this experiment, we fix $\beta_2$ as 10 and set the $\beta_1$ as 0.6, 0.7, 0.8, 0.9, 0.95 to check the influence of this parameter on the GOT-10k dataset. As shown in Table \ref{param_plpt}, the tracking results are relatively stable. Specifically, the AO/SR of these settings are 0.451/0.541, 0.451/0.541, 0.455/0.547, 0.456/0.549, 0.453/0.549. We can get the best performance when the $\beta_1$ is set as 0.8-0.9.

\begin{table}[!htp]
\center
\caption{Parameter analysis of $\beta_1$ and $\beta_2$ on the GOT-10k dataset.}  
\scriptsize 
\label{param_plpt}
\begin{tabular}{c|cccccc} 
\hline \toprule [0.8 pt]
$\beta_1$ & 4 &6 &8 &10 &12 \\
\hline 
AO|SR 	&0.450|0.543     &0.453|0.545	&0.455|0.548 &0.455|0.547  &0.454|0.545 \\
\hline \toprule [0.8 pt]
$\beta_2$ &0.6 &0.7 &0.8 &0.9 &0.95 \\
\hline 
AO|SR 	&0.451|0.541     &0.451|0.541	&0.455|0.547 &0.456|0.549  &0.453|0.549 \\
\hline \toprule [0.8 pt]
\end{tabular}
\end{table}

\begin{table}[!htp]
\center
\caption{Parameter analysis of $\beta_1$ and $\beta_2$ on the VOT2018LT dataset.}   
\small 
\label{param_plptvot2018lt}
\begin{tabular}{c|cccccc} 
\hline \toprule [0.8 pt]
$\beta_1$  &7 &9 &11 &13 &15 \\
\hline 
F1-score 	 &0.510    &0.507    &0.515    &0.522   &0.510 \\
\hline \toprule [0.8 pt]
$\beta_2$  &0.6    &0.7    &0.8    &0.85    &0.9     \\
\hline 
F1-score 	 &0.515    &0.516    &0.515    &0.515    &0.509     \\
\hline \toprule [0.8 pt]
\end{tabular}
\end{table}

Aforementioned analysis is mainly focused on the short-term tracking dataset, we also report the results on the long-term tracking dataset (the SiamMask is used to evaluate on the VOT2018LT dataset), as shown in Table \ref{param_plptvot2018lt}. We can find that the overall results will be better when the  $\beta_1$ and $\beta_2$ are set as 13 and 0.8, respectively. These results also demonstrate that our tracker is robust to these two parameters.

\begin{figure*}[!htb]
\center
\includegraphics[width=5.5in]{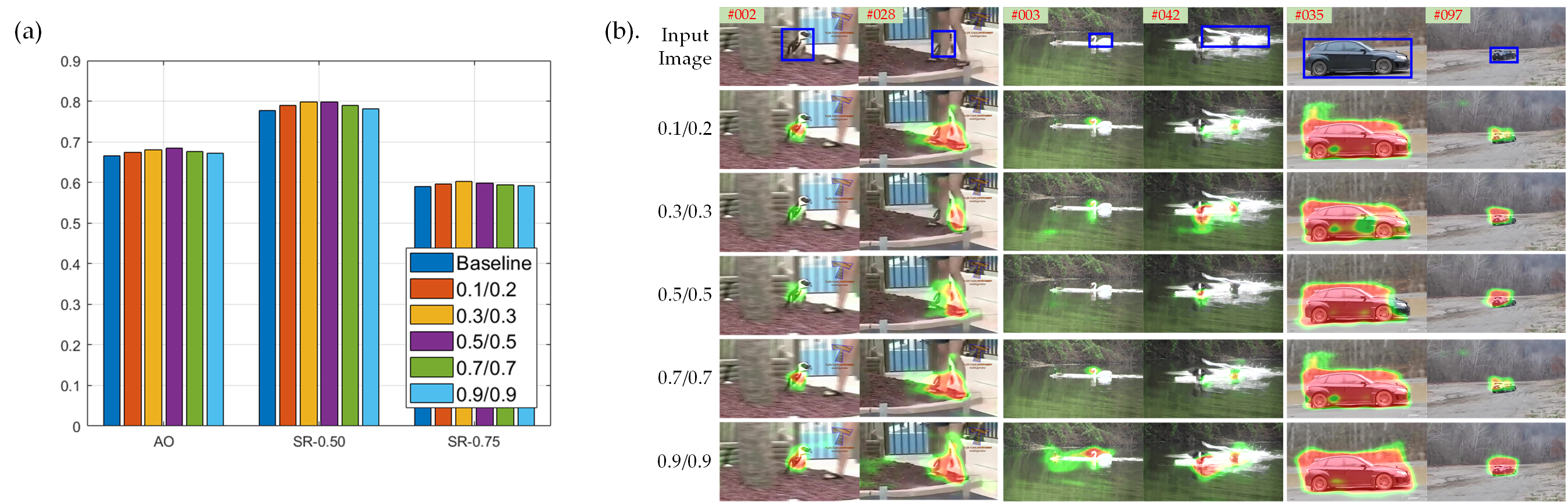}
\caption{(a). Tracking results on GOT-10k dataset with various $\lambda_1$ and $\lambda_2$;  (b). Attention maps predicted by various $\lambda_1/\lambda_2$.} 
\label{lambda_vis}
\end{figure*}

In addition, we also evaluate the tradeoff parameters $\lambda_1$ and $\lambda_2$ used in the adversarial learning phase. We integrate the TANet with DiMP tracker (SuperDiMP version is adopted in this experiment) and the detailed tracking results on GOT-10k can be found in Fig. \ref{lambda_vis} (a). Specifically, the baseline tracker attains 0.666/0.777/0.589 on $AO/SR_{0.50}/SR_{0.75}$, respectively. After integrating our TANet into the DiMP for joint local and global based tracking, the overall performance can be improved with various values of  $\lambda_1$ and $\lambda_2$, as shown in Fig. \ref{lambda_vis} (a). Specifically, we can improve baseline tracker to 0.674/0.790/0.596, 0.681/0.798/0.603, 0.684/0.799/0.599, 0.677/0.790/0.594, 0.671/0.782/0.591, when the parameters $\lambda_1/\lambda_2$ are set as 0.1/0.2, 0.3/0.3, 0.5/0.5, 0.7/0.7, 0.9/0.9, respectively. We can find that the best performance can be obtained when the two parameters are set as 0.3-0.5. We also give a visualization of the attained attention maps with these values in Fig. \ref{lambda_vis} (b). From the qualitative and quantitative analysis, we can find that our TANet is relatively robust to these parameters and all improve the baseline tracker.

\subsection{Generalization} \label{generalization} 
In this work, we integrate our proposed module with multiple trackers to validate its generalization, as shown in Fig. \ref{trackerpipeline}. Specifically, we test \emph{classification} based trackers RT-MDNet \cite{jung2018real}, \emph{anchor-based Siamese network} based trackers THOR \cite{THOR2019}, SiamRPN++ \cite{li2019siamrpn++}, SiamMask \cite{wang2019siammask}, \emph{anchor-free Siamese network} based tracker SiamFC++ \cite{xu2020siamfc++}, and \emph{discriminative learning Siamese network} based tracker DiMP \cite{Bhat_dimp}. We denote trackers combined with our module with (\cmark), as shown in Table \ref{genericResults}.

We can find that all the used baseline trackers can be improved by integrating our TANet, as demonstrated in Table \ref{genericResults}. Specifically, the THOR achieves 0.447/0.538 (AO/SR) on the GOT-10k and 0.593/0.339/0.432 on the VOT2018LT dataset, while we can improve them to 0.459/0.551 and 0.600/0.452/0.516. We also improve the SiamMask from 0.451/0.541 to 0.455/0.547 on the GOT-10k, 0.610/0.407/0.488 to 0.600/0.471/0.528 on the VOT2018LT dataset, respectively. The tracking results of SiamRPN++, SiamFC++, RT-MDNet, and DiMP also validate the effectiveness of our TANet. Therefore, we can conclude that our proposed TANet is a plug and play module that can be simply integrated with existing trackers to further improve their tracking results.

\begin{table*}[htp!]
\center
\scriptsize  
\caption{Tracking results of 5 baseline trackers and ours on GOT-10k and VOT2018-LT.} \label{genericResults}
\begin{tabular}{c|cc|cc|cc|cc|cc}
\hline \toprule [1 pt]
\textbf{GOT-10k} 	&{SiamFC++ }   &{SiamFC++(\cmark)}    &{THOR }      &{THOR(\cmark)}   &{SiamRPN++ }       &{SiamRPN++(\cmark)}  &{SiamMask }   &{SiamMask(\cmark)}  &DiMP  &DiMP(\cmark) \\  
AO 		            &0.595       &0.607     &0.447       &0.459   &0.453      &0.454       &0.451       &0.455     &0.666		&0.684		\\
SR		            &0.695       &0.737     &0.538       &0.551   &0.537      &0.539       &0.541       &0.547    &0.777		&0.799		 \\
\hline \toprule [1 pt]
\textbf{VOT2018-LT} 	&{RT-MDNet }   &{RT-MDNet(\cmark)}    &{THOR}      &{THOR(\cmark)}   &{SiamRPN++}       &{SiamRPN++(\cmark)}  &{SiamMask}   &{SiamMask(\cmark)}  &DiMP  &DiMP(\cmark)   \\  
Precision 		&0.432   	 &0.484      &0.593       &0.600  &0.646      &0.649       &0.610        &0.600     &0.636	&0.695			\\
Recall 			&0.276       &0.317     &0.339       &0.452   &0.419      &0.535       &0.407       &0.471     &0.426	&0.491		 \\
F1-score 		&0.337       &0.383     &0.432       &0.516   &0.508      &0.586       &0.488       &0.528     &0.510	&0.575		  \\
\hline \toprule [1 pt]
\end{tabular}
\end{table*}

\subsection{Visualization} 
In addition to the aforementioned quantitative analysis, we also give some visualization about the learned target-aware attention and tracking results on multiple videos for qualitative analysis. 

\begin{figure}[!htb]
\center
\includegraphics[width=3.5in]{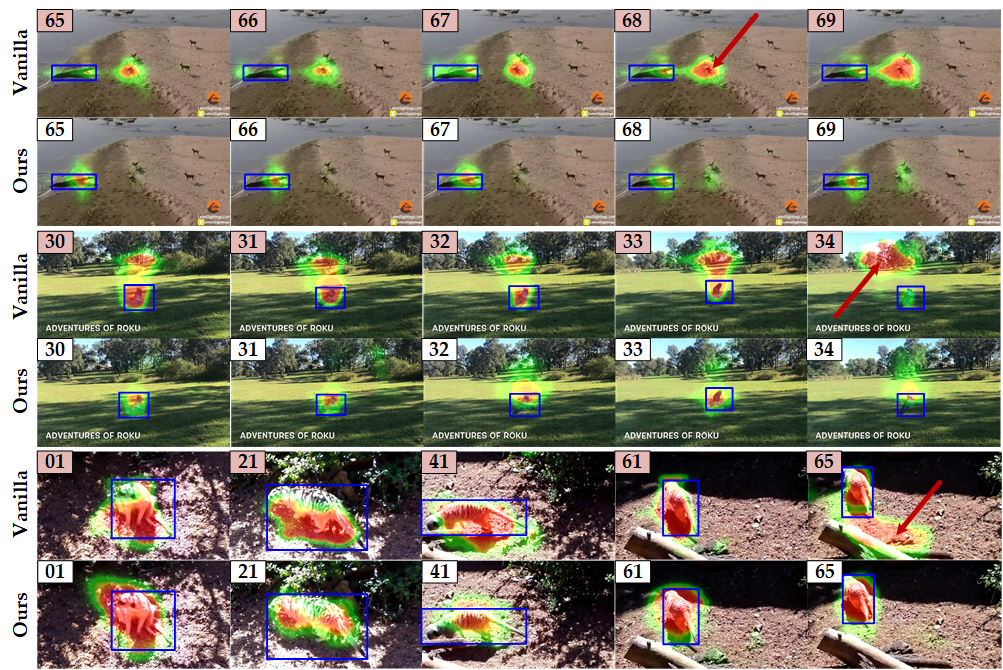}
\caption{Attention map predicted by vanilla and our adversarial learning based TANet.} 
\label{ganTANet_vis}
\end{figure}

\textbf{Target-Aware Attention.}  
As shown in Fig. \ref{ganTANet_vis}, we can find that vanilla attention maps predicted by Algorithm-1 (i.e., only ResNet and BCE loss function used) are weak at target object regions in some videos, like \emph{crocodile} in the first row\footnote{The target object is denoted with blue rectangles.}. Meanwhile, the background in these maps are somewhat have a higher response score. Similar cases can also be found in the third and fifth row (We use a couple of red arrows to highlight these regions). These noisy regions will make the trackers tend to model drift when using global search. Compared with these attention maps, our adversarial learning based generator can predict more accurate candidate search regions. Our attention maps are more consistent from the temporal view. Therefore, we can achieve better tracking results compared with baseline local search trackers.

\begin{figure}[!htb]
\center
\includegraphics[width=3.5in]{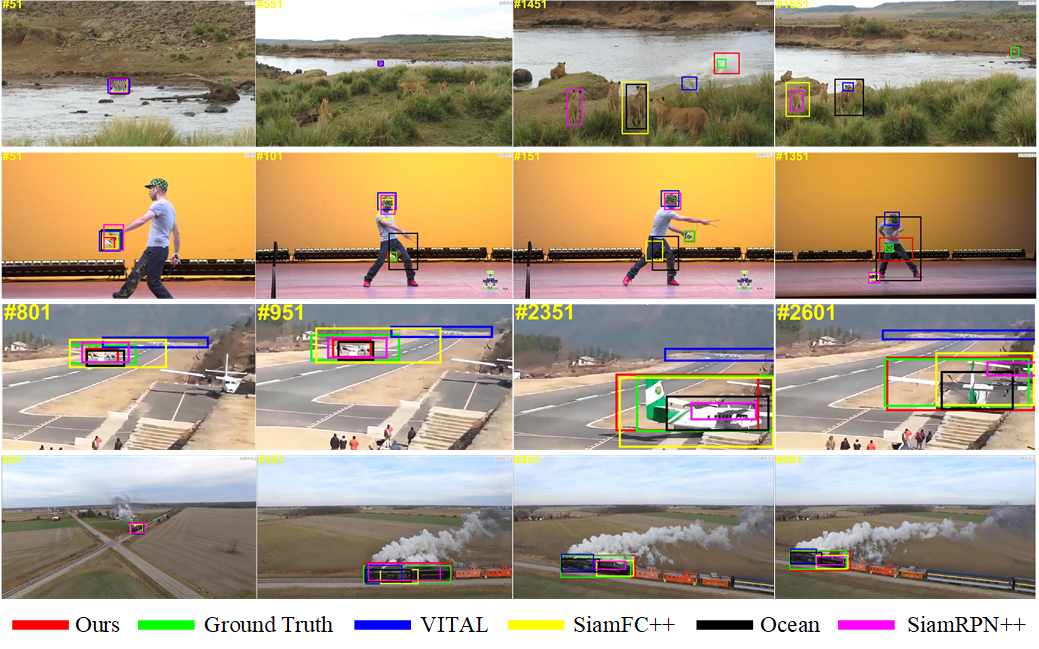}
\caption{Comparison between our method and other SOTA trackers on the LaSOT dataset.} 
\label{vis_trackingresults}
\end{figure}

\textbf{Tracking Results.}
We give a visualization of the tracking results of our tracker (SiamFC++ based) and other compared state-of-the-art (SOTA) algorithms in Fig. \ref{vis_trackingresults}. It is easy to find that our tracker is more robust to fast motion, heavy occlusion, and scale variation, etc. These experimental results all validate the effectiveness and advantages of our proposed TANet for global search.

\subsection{Discussion}
Tracking with fast motion, heavy occlusion, deformation occurred have been explored in many state-of-the-art trackers. For the fast motion, current trackers usually depends on strong feature representation of deep learning \cite{yang2020roam, voigtlaender2020siamRCNN, lukezic2020d3s, danelljan2019atom}, or pre-processing operation like de-blur \cite{bai2020deepDeblur}, high frame rate camera \cite{kiani2017nfs}. Some trackers also adopt hard sample generation for the motion blur caused by fast motion or heavy occlusion \cite{Wang_2018_CVPR, SongYiBing_2018_CVPR}. For the deformation, they usually employ graph matching \cite{wang2019deformable}, gated fusion \cite{liu2019deformable} for robust tracking. These works can handle specific challenge from aforementioned views, however, seldom of them can address all of them. Therefore, they may still facing model drift and how to re-capture the target object as soon as possible when these challenges passed away is the main consideration of our work. In contrast, our proposed target-aware attention model focus on providing candidate search regions from global view. Because we observe that these challenging situations only occupies a small part of the whole video sequences. This inspires us to think why don’t we design an global search module to re-capture the target object once the challenging frames passed away? This is the main motivation of this work that conduct tracking in a joint local and global search manner. Thus, we design a novel target-aware attention model based on generative adversarial learning framework and our experimental results on multiple tracking benchmarks all validated the effectiveness, flexibility, and efficiency of our tracker.

\subsection{Failed Cases} 
\begin{figure}[!htb]
\center
\includegraphics[width=3.5in]{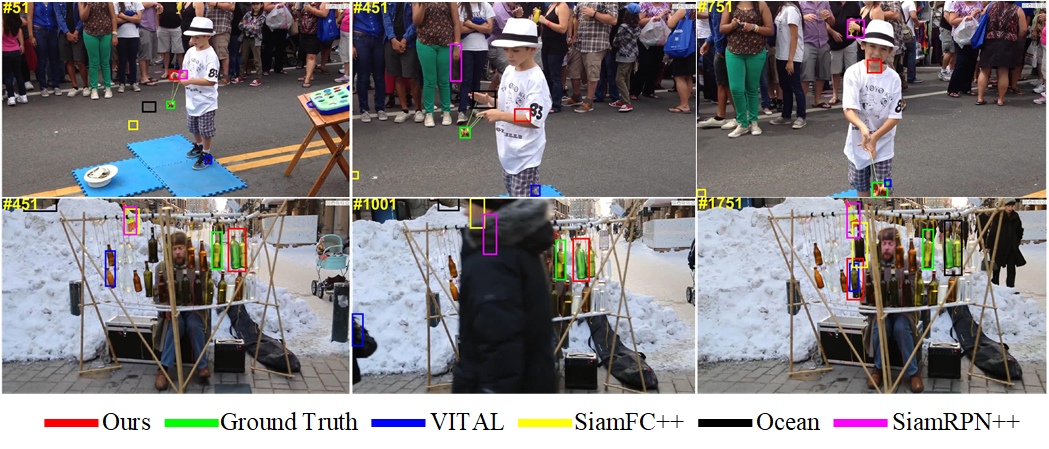}
\caption{Failed cases of our tracker and other state of the art trackers.} 
\label{failedCases}
\end{figure}

According to the aforementioned analysis, we can find that our proposed target-aware attention network can provide candidate regions for global search and it works well on tracking task especially the long-term tracking task. At the same time, we simply switch local and global search module according to the given threshold which can't handle videos which already failed to locate the target object, but the predicted scores are high. Besides, our tracker is also easily influenced by similar small targets, and we give some visualization of the failed cases in Fig. \ref{failedCases}. The \emph{yoyo} in the boy's hand, the \emph{bottle} in the second column are hard to track due to the clutter background.

\section{Conclusion and Future Works}
In this paper, we propose a novel target-aware attention prediction network for global search in the single object tracking task. Specifically, we use 2D and 3D convolutional neural network for spatial and temporal information encoding. More importantly, we introduce the target template as the condition to predict target driven attention maps. These features are concatenated together and fed into an upsample network to generate corresponding attention maps recurrently. In addition to standard BCE loss used for the training, we also introduce the adversarial learning for more accurate attention generation. The appearance and motion discriminators are introduced to model the relations between nearby pixels and temporal consistency between continuous video frames. Extensive experiments on multiple trackers and multiple benchmark datasets all validated the effectiveness, generalization, and advantages of our proposed attention for tracking task.

In our future works, we will consider designing a novel adaptive switch network to replace the naive score-based switch scheme used in this paper. Besides, we will also consider exploring the global structured scene information for robust tracking.

\section*{Acknowledgements}	
This work is jointly supported by Key-Area Research and Development Program of Guangdong Province 2019B010155002, Postdoctoral Innovative Talent Support Program BX20200174, China Postdoctoral Science Foundation Funded Project 2020M682828, National Nature Science Foundation of China (61860206004, 61825101, 62076003), and the Joint Funds of the National Natural Science Foundation of China (Grant NO. U20B2052).

{
\bibliographystyle{IEEEtran}
\bibliography{reference}

\begin{thebibliography}{10}
\providecommand{\url}[1]{#1}
\csname url@samestyle\endcsname
\providecommand{\newblock}{\relax}
\providecommand{\bibinfo}[2]{#2}
\providecommand{\BIBentrySTDinterwordspacing}{\spaceskip=0pt\relax}
\providecommand{\BIBentryALTinterwordstretchfactor}{4}
\providecommand{\BIBentryALTinterwordspacing}{\spaceskip=\fontdimen2\font plus
\BIBentryALTinterwordstretchfactor\fontdimen3\font minus
  \fontdimen4\font\relax}
\providecommand{\BIBforeignlanguage}[2]{{%
\expandafter\ifx\csname l@#1\endcsname\relax
\typeout{** WARNING: IEEEtran.bst: No hyphenation pattern has been}%
\typeout{** loaded for the language `#1'. Using the pattern for}%
\typeout{** the default language instead.}%
\else
\language=\csname l@#1\endcsname
\fi
#2}}
\providecommand{\BIBdecl}{\relax}
\BIBdecl

\bibitem{li2018siamRPN}
B.~Li, J.~Yan, W.~Wu, Z.~Zhu, and X.~Hu, ``High performance visual tracking
  with siamese region proposal network,'' in \emph{Proceedings of the IEEE
  Conference on Computer Vision and Pattern Recognition}, 2018, pp. 8971--8980.

\bibitem{song2017crest}
Y.~Song, C.~Ma, L.~Gong, J.~Zhang, R.~W. Lau, and M.-H. Yang, ``Crest:
  Convolutional residual learning for visual tracking,'' in \emph{Proceedings
  of the IEEE International Conference on Computer Vision}, 2017, pp.
  2555--2564.

\bibitem{Nam2015Learning}
H.~Nam and B.~Han, ``Learning multi-domain convolutional neural networks for
  visual tracking,'' in \emph{Proceedings of the IEEE Conference on Computer
  Vision and Pattern Recognition}, 2016, pp. 4293--4302.

\bibitem{Danelljan2016ECO}
M.~Danelljan, G.~Bhat, F.~Shahbaz~Khan, and M.~Felsberg, ``Eco: Efficient
  convolution operators for tracking,'' in \emph{Proceedings of the IEEE
  conference on computer vision and pattern recognition}, 2017, pp. 6638--6646.

\bibitem{Danelljan2016Beyond}
M.~Danelljan, A.~Robinson, F.~S. Khan, and M.~Felsberg, ``Beyond correlation
  filters: Learning continuous convolution operators for visual tracking,'' in
  \emph{European conference on computer vision}.\hskip 1em plus 0.5em minus
  0.4em\relax Springer, 2016, pp. 472--488.

\bibitem{hu2017manifold}
H.~Hu, B.~Ma, J.~Shen, and L.~Shao, ``Manifold regularized correlation object
  tracking,'' \emph{IEEE transactions on neural networks and learning systems},
  vol.~29, no.~5, pp. 1786--1795, 2017.

\bibitem{zhu2018dynamic}
G.~Zhu, Z.~Zhang, J.~Wang, Y.~Wu, and H.~Lu, ``Dynamic collaborative
  tracking,'' \emph{IEEE Transactions on Neural Networks and Learning Systems},
  vol.~30, no.~10, pp. 3035--3046, 2018.

\bibitem{ge2020cascaded}
S.~{Ge}, C.~{Zhang}, S.~{Li}, D.~{Zeng}, and D.~{Tao}, ``Cascaded correlation
  refinement for robust deep tracking,'' \emph{IEEE Transactions on Neural
  Networks and Learning Systems}, vol.~32, no.~3, pp. 1276--1288, 2021.

\bibitem{yun2018action}
S.~Yun, J.~Choi, Y.~Yoo, K.~Yun, and J.~Y. Choi, ``Action-driven visual object
  tracking with deep reinforcement learning,'' \emph{IEEE transactions on
  neural networks and learning systems}, vol.~29, no.~6, pp. 2239--2252, 2018.

\bibitem{liwicki2012efficient}
S.~Liwicki, S.~Zafeiriou, G.~Tzimiropoulos, and M.~Pantic, ``Efficient online
  subspace learning with an indefinite kernel for visual tracking and
  recognition,'' \emph{IEEE transactions on neural networks and learning
  systems}, vol.~23, no.~10, pp. 1624--1636, 2012.

\bibitem{wang2021tnl2k}
W.~Xiao, S.~Xiujun, Z.~Zhipeng, J.~Bo, W.~Yaowei, T.~Yonghong, and W.~Feng,
  ``Towards more flexible and accurate object tracking with natural language:
  Algorithms and benchmark,'' in \emph{The IEEE/CVF Conference on Computer
  Vision and Pattern Recognition (CVPR)}, 2021.

\bibitem{zhang2019deeper}
Z.~Zhang and H.~Peng, ``Deeper and wider siamese networks for real-time visual
  tracking,'' in \emph{Proceedings of the IEEE Conference on Computer Vision
  and Pattern Recognition}, 2019, pp. 4591--4600.

\bibitem{li2019siamrpn++}
B.~Li, W.~Wu, Q.~Wang, F.~Zhang, J.~Xing, and J.~Yan, ``Siamrpn++: Evolution of
  siamese visual tracking with very deep networks,'' in \emph{Proceedings of
  the IEEE Conference on Computer Vision and Pattern Recognition}, 2019, pp.
  4282--4291.

\bibitem{huang2019globaltrack}
L.~Huang, X.~Zhao, and K.~Huang, ``Globaltrack: A simple and strong baseline
  for long-term tracking,'' in \emph{Proceedings of the AAAI Conference on
  Artificial Intelligence}, vol.~34, no.~07, 2020, pp. 11\,037--11\,044.

\bibitem{zhu2016beyond}
G.~Zhu, F.~Porikli, and H.~Li, ``Beyond local search: Tracking objects
  everywhere with instance-specific proposals,'' in \emph{Proceedings of the
  IEEE Conference on Computer Vision and Pattern Recognition}, 2016, pp.
  943--951.

\bibitem{zhang2018learningTermTracking}
Y.~Zhang, D.~Wang, L.~Wang, J.~Qi, and H.~Lu, ``Learning regression and
  verification networks for long-term visual tracking,'' \emph{arXiv preprint
  arXiv:1809.04320}, 2018.

\bibitem{yan2019skimming}
B.~Yan, H.~Zhao, D.~Wang, H.~Lu, and X.~Yang, ``'skimming-perusal'tracking: A
  framework for real-time and robust long-term tracking,'' in \emph{Proceedings
  of the IEEE International Conference on Computer Vision}, 2019, pp.
  2385--2393.

\bibitem{long2015fullyFCNsegmentation}
J.~Long, E.~Shelhamer, and T.~Darrell, ``Fully convolutional networks for
  semantic segmentation,'' in \emph{Proceedings of the IEEE conference on
  computer vision and pattern recognition}, 2015, pp. 3431--3440.

\bibitem{chen2018deeplab}
L.-C. Chen, G.~Papandreou, I.~Kokkinos, K.~Murphy, and A.~L. Yuille, ``Deeplab:
  Semantic image segmentation with deep convolutional nets, atrous convolution,
  and fully connected crfs,'' \emph{IEEE transactions on pattern analysis and
  machine intelligence}, vol.~40, no.~4, pp. 834--848, 2018.

\bibitem{gong2017lookintoPerson}
K.~Gong, X.~Liang, D.~Zhang, X.~Shen, and L.~Lin, ``Look into person:
  Self-supervised structure-sensitive learning and a new benchmark for human
  parsing,'' in \emph{Proceedings of the IEEE Conference on Computer Vision and
  Pattern Recognition}, 2017, pp. 932--940.

\bibitem{liang2015deepParse}
X.~Liang, S.~Liu, X.~Shen, J.~Yang, L.~Liu, J.~Dong, L.~Lin, and S.~Yan, ``Deep
  human parsing with active template regression,'' \emph{IEEE transactions on
  pattern analysis and machine intelligence}, vol.~37, no.~12, pp. 2402--2414,
  2015.

\bibitem{luo2018macroAdversarial}
Y.~Luo, Z.~Zheng, L.~Zheng, T.~Guan, J.~Yu, and Y.~Yang, ``Macro-micro
  adversarial network for human parsing,'' in \emph{Proceedings of the European
  Conference on Computer Vision (ECCV)}, 2018, pp. 418--434.

\bibitem{luc2016semanticGAN}
P.~Luc, C.~Couprie, S.~Chintala, and J.~Verbeek, ``Semantic segmentation using
  adversarial networks,'' \emph{arXiv preprint arXiv:1611.08408}, 2016.

\bibitem{tang2019salientDetection}
Y.~Tang and X.~Wu, ``Salient object detection using cascaded convolutional
  neural networks and adversarial learning,'' \emph{IEEE Transactions on
  Multimedia}, vol.~21, no.~9, pp. 2237--2247, 2019.

\bibitem{wang2019qualityRGBTsaliency}
X.~Wang, T.~Sun, R.~Yang, C.~Li, B.~Luo, and J.~Tang, ``Quality-aware
  dual-modal saliency detection via deep reinforcement learning,'' \emph{Signal
  Processing: Image Communication}, vol.~75, pp. 158--167, 2019.

\bibitem{nguyen2017shadowDetectionGAN}
V.~Nguyen, Y.~Vicente, F.~Tomas, M.~Zhao, M.~Hoai, and D.~Samaras, ``Shadow
  detection with conditional generative adversarial networks,'' in
  \emph{Proceedings of the IEEE International Conference on Computer Vision},
  2017, pp. 4510--4518.

\bibitem{wang2018stackedDetectionGAN}
J.~Wang, X.~Li, and J.~Yang, ``Stacked conditional generative adversarial
  networks for jointly learning shadow detection and shadow removal,'' in
  \emph{Proceedings of the IEEE Conference on Computer Vision and Pattern
  Recognition}, 2018, pp. 1788--1797.

\bibitem{wang2019GANTrack}
X.~Wang, R.~Yang, T.~Sun, and B.~Luo, ``Learning target-aware attention for
  robust tracking with conditional adversarial network.'' in \emph{30TH British
  Machine Vision Conference}, 2019, p. 131.

\bibitem{li2018TrackSurvey}
P.~Li, D.~Wang, L.~Wang, and H.~Lu, ``Deep visual tracking: Review and
  experimental comparison,'' \emph{Pattern Recognition}, vol.~76, pp. 323--338,
  2018.

\bibitem{marvasti2021TrackSurvey}
S.~M. {Marvasti-Zadeh}, L.~{Cheng}, H.~{Ghanei-Yakhdan}, and S.~{Kasaei},
  ``Deep learning for visual tracking: A comprehensive survey,'' \emph{IEEE
  Transactions on Intelligent Transportation Systems}, pp. 1--26, 2021.

\bibitem{brunetti2018CVDLSurvey}
A.~Brunetti, D.~Buongiorno, G.~F. Trotta, and V.~Bevilacqua, ``Computer vision
  and deep learning techniques for pedestrian detection and tracking: A
  survey,'' \emph{Neurocomputing}, vol. 300, pp. 17--33, 2018.

\bibitem{han2017branchout}
B.~Han, J.~Sim, and H.~Adam, ``Branchout: Regularization for online ensemble
  tracking with convolutional neural networks,'' in \emph{Proceedings of IEEE
  International Conference on Computer Vision}, 2017, pp. 2217--2224.

\bibitem{Park_2018_ECCV}
E.~Park and A.~C. Berg, ``Meta-tracker: Fast and robust online adaptation for
  visual object trackers,'' in \emph{Proceedings of the European Conference on
  Computer Vision (ECCV)}, 2018, pp. 569--585.

\bibitem{jung2018real}
I.~Jung, J.~Son, M.~Baek, and B.~Han, ``Real-time mdnet,'' in \emph{Proceedings
  of the European Conference on Computer Vision (ECCV)}, 2018, pp. 83--98.

\bibitem{SongYiBing_2018_CVPR}
Y.~Song, C.~Ma, X.~Wu, L.~Gong, L.~Bao, W.~Zuo, C.~Shen, R.~W. Lau, and M.-H.
  Yang, ``Vital: Visual tracking via adversarial learning,'' in
  \emph{Proceedings of the IEEE conference on computer vision and pattern
  recognition}, 2018, pp. 8990--8999.

\bibitem{Li2017ReGLe}
C.~Li, X.~Wu, Z.~Bao, and J.~Tang, ``Regle: spatially regularized graph
  learning for visual tracking,'' in \emph{Proceedings of the 25th ACM
  international conference on Multimedia}, 2017, pp. 252--260.

\bibitem{hare2015struck}
S.~Hare, S.~Golodetz, A.~Saffari, V.~Vineet, M.-M. Cheng, S.~L. Hicks, and
  P.~H. Torr, ``Struck: Structured output tracking with kernels,'' \emph{IEEE
  transactions on pattern analysis and machine intelligence}, vol.~38, no.~10,
  pp. 2096--2109, 2015.

\bibitem{danelljan2015learning}
M.~Danelljan, G.~Hager, F.~Shahbaz~Khan, and M.~Felsberg, ``Learning spatially
  regularized correlation filters for visual tracking,'' in \emph{Proceedings
  of the IEEE International Conference on Computer Vision}, 2015, pp.
  4310--4318.

\bibitem{choi2017attentional}
J.~Choi, H.~Jin~Chang, S.~Yun, T.~Fischer, Y.~Demiris, and J.~Young~Choi,
  ``Attentional correlation filter network for adaptive visual tracking,'' in
  \emph{Proceedings of the IEEE conference on computer vision and pattern
  recognition}, 2017, pp. 4807--4816.

\bibitem{Bolme2010Visual}
D.~S. Bolme, J.~R. Beveridge, B.~A. Draper, and Y.~M. Lui, ``Visual object
  tracking using adaptive correlation filters,'' vol. 119, no.~5, pp.
  2544--2550, 2010.

\bibitem{valmadre2017end}
J.~Valmadre, L.~Bertinetto, J.~Henriques, A.~Vedaldi, and P.~H. Torr,
  ``End-to-end representation learning for correlation filter based tracking,''
  in \emph{Computer Vision and Pattern Recognition (CVPR), 2017 IEEE Conference
  on}.\hskip 1em plus 0.5em minus 0.4em\relax IEEE, 2017, pp. 5000--5008.

\bibitem{lukezic2017discriminative}
A.~Lukezic, T.~Vojir, L.~C. Zajc, J.~Matas, and M.~Kristan, ``Discriminative
  correlation filter with channel and spatial reliability,'' in \emph{2017 IEEE
  Conference on Computer Vision and Pattern Recognition (CVPR)}.\hskip 1em plus
  0.5em minus 0.4em\relax IEEE, 2017, pp. 4847--4856.

\bibitem{Henriques2015High}
J.~F. Henriques, R.~Caseiro, P.~Martins, and J.~Batista, ``High-speed tracking
  with kernelized correlation filters,'' \emph{IEEE Transactions on Pattern
  Analysis Machine Intelligence}, vol.~37, no.~3, pp. 583--596, 2015.

\bibitem{wang2005svm}
T.~Evgeniou and M.~Pontil, ``Support vector machines: Theory and
  applications,'' in \emph{Advanced Course on Artificial Intelligence}.\hskip
  1em plus 0.5em minus 0.4em\relax Springer, 1999, pp. 249--257.

\bibitem{lecun2015DLnature}
Y.~LeCun, Y.~Bengio, and G.~Hinton, ``Deep learning,'' \emph{nature}, vol. 521,
  no. 7553, pp. 436--444, 2015.

\bibitem{fan2019lasot}
H.~Fan, L.~Lin, F.~Yang, P.~Chu, G.~Deng, S.~Yu, H.~Bai, Y.~Xu, C.~Liao, and
  H.~Ling, ``Lasot: A high-quality benchmark for large-scale single object
  tracking,'' in \emph{Proceedings of the IEEE Conference on Computer Vision
  and Pattern Recognition}, 2019, pp. 5374--5383.

\bibitem{muller2018trackingnet}
M.~Muller, A.~Bibi, S.~Giancola, S.~Alsubaihi, and B.~Ghanem, ``Trackingnet: A
  large-scale dataset and benchmark for object tracking in the wild,'' in
  \emph{Proceedings of the European Conference on Computer Vision (ECCV)},
  2018, pp. 300--317.

\bibitem{bertinetto2016fully}
L.~Bertinetto, J.~Valmadre, J.~F. Henriques, A.~Vedaldi, and P.~H. Torr,
  ``Fully-convolutional siamese networks for object tracking,'' in
  \emph{European conference on computer vision}.\hskip 1em plus 0.5em minus
  0.4em\relax Springer, 2016, pp. 850--865.

\bibitem{Tao2016Siamese}
R.~Tao, E.~Gavves, and A.~W. Smeulders, ``Siamese instance search for
  tracking,'' in \emph{Proceedings of the IEEE Conference on Computer Vision
  and Pattern Recognition}, 2016, pp. 1420--1429.

\bibitem{held2016goturn}
D.~Held, S.~Thrun, and S.~Savarese, ``Learning to track at 100 fps with deep
  regression networks,'' in \emph{European Conference on Computer
  Vision}.\hskip 1em plus 0.5em minus 0.4em\relax Springer, 2016, pp. 749--765.

\bibitem{zhang2019siamDW}
Z.~Zhang and H.~Peng, ``Deeper and wider siamese networks for real-time visual
  tracking,'' in \emph{Proceedings of the IEEE Conference on Computer Vision
  and Pattern Recognition}, 2019, pp. 4591--4600.

\bibitem{xu2020siamfc++}
Y.~Xu, Z.~Wang, Z.~Li, Y.~Yuan, and G.~Yu, ``Siamfc++: Towards robust and
  accurate visual tracking with target estimation guidelines.'' in \emph{AAAI},
  2020, pp. 12\,549--12\,556.

\bibitem{Wang_2018_CVPR}
X.~Wang, C.~Li, B.~Luo, and J.~Tang, ``Sint++: Robust visual tracking via
  adversarial positive instance generation,'' in \emph{Proceedings of the IEEE
  conference on computer vision and pattern recognition}, 2018, pp. 4864--4873.

\bibitem{zhao2018adversarial}
F.~Zhao, J.~Wang, Y.~Wu, and M.~Tang, ``Adversarial deep tracking,'' \emph{IEEE
  Transactions on Circuits and Systems for Video Technology}, vol.~29, no.~7,
  pp. 1998--2011, 2018.

\bibitem{danelljan2019atom}
M.~Danelljan, G.~Bhat, F.~S. Khan, and M.~Felsberg, ``Atom: Accurate tracking
  by overlap maximization,'' in \emph{Proceedings of the IEEE Conference on
  Computer Vision and Pattern Recognition}, 2019, pp. 4660--4669.

\bibitem{Bhat_dimp}
G.~Bhat, M.~Danelljan, L.~V. Gool, and R.~Timofte, ``Learning discriminative
  model prediction for tracking,'' in \emph{Proceedings of the IEEE/CVF
  International Conference on Computer Vision}, 2019, pp. 6182--6191.

\bibitem{danelljan2020prDIMP}
M.~Danelljan, L.~V. Gool, and R.~Timofte, ``Probabilistic regression for visual
  tracking,'' in \emph{Proceedings of the IEEE/CVF Conference on Computer
  Vision and Pattern Recognition}, 2020, pp. 7183--7192.

\bibitem{lukezic2020d3s}
A.~Lukezic, J.~Matas, and M.~Kristan, ``D3s-a discriminative single shot
  segmentation tracker,'' in \emph{Proceedings of the IEEE/CVF Conference on
  Computer Vision and Pattern Recognition}, 2020, pp. 7133--7142.

\bibitem{fang2019LGTrack}
Y.~Fang, S.~Ko, and G.-S. Jo, ``Robust visual tracking based on
  global-and-local search with confidence reliability estimation,''
  \emph{Neurocomputing}, vol. 367, pp. 273--286, 2019.

\bibitem{ma2015LTCF}
C.~Ma, X.~Yang, C.~Zhang, and M.-H. Yang, ``Long-term correlation tracking,''
  in \emph{Proceedings of the IEEE conference on computer vision and pattern
  recognition}, 2015, pp. 5388--5396.

\bibitem{ramesh2018LTEvent}
B.~Ramesh, S.~Zhang, Z.~W. Lee, Z.~Gao, G.~Orchard, and C.~Xiang, ``Long-term
  object tracking with a moving event camera.'' in \emph{Bmvc}, 2018, p. 241.

\bibitem{voigtlaender2019mots}
P.~Voigtlaender, M.~Krause, A.~Osep, J.~Luiten, B.~B.~G. Sekar, A.~Geiger, and
  B.~Leibe, ``Mots: Multi-object tracking and segmentation,'' in
  \emph{Proceedings of the IEEE/CVF Conference on Computer Vision and Pattern
  Recognition}, 2019, pp. 7942--7951.

\bibitem{zhang2020TRL}
W.~Zhang, R.~Song, Y.~Li \emph{et~al.}, ``Online decision based visual tracking
  via reinforcement learning,'' \emph{Advances in Neural Information Processing
  Systems}, vol.~33, 2020.

\bibitem{xuan2021siamLT}
S.~Xuan, S.~Li, Z.~Zhao, L.~Kou, Z.~Zhou, and G.-S. Xia, ``Siamese networks
  with distractor-reduction method for long-term visual object tracking,''
  \emph{Pattern Recognition}, vol. 112, p. 107698, 2021.

\bibitem{dai2020high}
K.~Dai, Y.~Zhang, D.~Wang, J.~Li, H.~Lu, and X.~Yang, ``High-performance
  long-term tracking with meta-updater,'' in \emph{Proceedings of the IEEE/CVF
  Conference on Computer Vision and Pattern Recognition}, 2020, pp. 6298--6307.

\bibitem{voigtlaender2020siamRCNN}
P.~Voigtlaender, J.~Luiten, P.~H. Torr, and B.~Leibe, ``Siam r-cnn: Visual
  tracking by re-detection,'' in \emph{Proceedings of the IEEE/CVF Conference
  on Computer Vision and Pattern Recognition}, 2020, pp. 6578--6588.

\bibitem{Choi2016Visual}
J.~Choi, H.~J. Chang, J.~Jeong, Y.~Demiris, and J.~Y. Choi, ``Visual tracking
  using attention-modulated disintegration and integration,'' in
  \emph{Proceedings of the IEEE conference on computer vision and pattern
  recognition}, 2016, pp. 4321--4330.

\bibitem{Hong2015Online}
S.~Hong, T.~You, S.~Kwak, and B.~Han, ``Online tracking by learning
  discriminative saliency map with convolutional neural network,'' in
  \emph{International Conference on Machine Learning}, 2015, pp. 597--606.

\bibitem{he2016KPSRT}
Z.~He, S.~Yi, Y.-M. Cheung, X.~You, and Y.~Y. Tang, ``Robust object tracking
  via key patch sparse representation,'' \emph{IEEE transactions on
  cybernetics}, vol.~47, no.~2, pp. 354--364, 2016.

\bibitem{wang2021deepmta}
X.~Wang, Z.~Chen, J.~Tang, B.~Luo, Y.~Wang, Y.~Tian, and F.~Wu, ``Dynamic
  attention guided multi-trajectory analysis for single object tracking,''
  \emph{IEEE Transactions on Circuits and Systems for Video Technology}, 2021.

\bibitem{chu2017online}
Q.~Chu, W.~Ouyang, H.~Li, X.~Wang, B.~Liu, and N.~Yu, ``Online multi-object
  tracking using cnn-based single object tracker with spatial-temporal
  attention mechanism,'' in \emph{2017 IEEE International Conference on
  Computer Vision (ICCV).(Oct 2017)}, 2017, pp. 4846--4855.

\bibitem{yang2018dmn}
T.~Yang and A.~B. Chan, ``Learning dynamic memory networks for object
  tracking,'' in \emph{Proceedings of the European Conference on Computer
  Vision (ECCV)}, 2018, pp. 152--167.

\bibitem{li2020robustTrack}
D.~Li, G.~Wen, Y.~Kuai, L.~Zhu, and F.~Porikli, ``Robust visual tracking with
  channel attention and focal loss,'' \emph{Neurocomputing}, vol. 401, pp.
  295--307, 2020.

\bibitem{Goodfellow2014Generative}
I.~J. Goodfellow, J.~Pouget-Abadie, M.~Mirza, B.~Xu, D.~Warde-Farley, S.~Ozair,
  A.~Courville, and Y.~Bengio, ``Generative adversarial nets,'' in
  \emph{International Conference on Neural Information Processing Systems},
  2014, pp. 2672--2680.

\bibitem{zhao2020antidecay}
F.~{Zhao}, T.~{Zhang}, Y.~{Wu}, M.~{Tang}, and J.~{Wang}, ``Antidecay lstm for
  siamese tracking with adversarial learning,'' \emph{IEEE Transactions on
  Neural Networks and Learning Systems}, pp. 1--15, 2020.

\bibitem{tran2018closer3DCNN}
D.~Tran, H.~Wang, L.~Torresani, J.~Ray, Y.~LeCun, and M.~Paluri, ``A closer
  look at spatiotemporal convolutions for action recognition,'' in
  \emph{Proceedings of the IEEE conference on Computer Vision and Pattern
  Recognition}, 2018, pp. 6450--6459.

\bibitem{he2016identityResNet}
K.~He, X.~Zhang, S.~Ren, and J.~Sun, ``Identity mappings in deep residual
  networks,'' in \emph{European conference on computer vision}.\hskip 1em plus
  0.5em minus 0.4em\relax Springer, 2016, pp. 630--645.

\bibitem{Pan2017SalGAN}
J.~Pan, E.~Sayrol, X.~G.-i. Nieto, C.~C. Ferrer, J.~Torres, K.~McGuinness, and
  N.~E. OConnor, ``Salgan: Visual saliency prediction with adversarial
  networks,'' in \emph{CVPR Scene Understanding Workshop (SUNw)}, 2017.

\bibitem{JifengWang_2018_CVPR}
J.~Wang, X.~Li, and J.~Yang, ``Stacked conditional generative adversarial
  networks for jointly learning shadow detection and shadow removal,'' in
  \emph{Proceedings of the IEEE Conference on Computer Vision and Pattern
  Recognition}, 2018, pp. 1788--1797.

\bibitem{THOR2019}
A.~Sauer, E.~Aljalbout, and S.~Haddadin, ``Tracking holistic object
  representations,'' in \emph{British Machine Vision Conference (BMVC)}, 2019.

\bibitem{wang2019siammask}
Q.~Wang, L.~Zhang, L.~Bertinetto, W.~Hu, and P.~H. Torr, ``Fast online object
  tracking and segmentation: A unifying approach,'' in \emph{Proceedings of the
  IEEE conference on computer vision and pattern recognition}, 2019, pp.
  1328--1338.

\bibitem{wu2015object}
Y.~Wu, J.~Lim, and M.-H. Yang, ``Object tracking benchmark,'' \emph{IEEE
  Transactions on Pattern Analysis and Machine Intelligence}, vol.~37, no.~9,
  pp. 1834--1848, 2015.

\bibitem{huang2019got}
L.~{Huang}, X.~{Zhao}, and K.~{Huang}, ``Got-10k: A large high-diversity
  benchmark for generic object tracking in the wild,'' \emph{IEEE Transactions
  on Pattern Analysis and Machine Intelligence}, vol.~43, no.~5, pp.
  1562--1577, 2021.

\bibitem{vot2018}
M.~Kristan, A.~Leonardis, J.~Matas, M.~Felsberg, R.~Pfugfelder, L.~\v{C}ehovin
  Zajc, T.~Vojir, G.~Bhat, A.~Lukezic, A.~Eldesokey, G.~Fernandez, and et~al.,
  ``The sixth visual object tracking vot2018 challenge results,'' 2018.

\bibitem{kristanvot2019}
M.~Kristan, J.~Matas, A.~Leonardis, M.~Felsberg, R.~Pflugfelder, J.-K.
  Kamarainen, L.~Cehovin~Zajc, O.~Drbohlav, A.~Lukezic, A.~Berg \emph{et~al.},
  ``The seventh visual object tracking vot2019 challenge results,'' in
  \emph{Proceedings of the IEEE International Conference on Computer Vision
  Workshops}, 2019.

\bibitem{duchi2011adaptive}
J.~Duchi, E.~Hazan, and Y.~Singer, ``Adaptive subgradient methods for online
  learning and stochastic optimization.'' \emph{Journal of machine learning
  research}, vol.~12, no.~7, 2011.

\bibitem{paszke2019pytorch}
A.~Paszke, S.~Gross, F.~Massa, A.~Lerer, J.~Bradbury, G.~Chanan, T.~Killeen,
  Z.~Lin, N.~Gimelshein, L.~Antiga \emph{et~al.}, ``Pytorch: An imperative
  style, high-performance deep learning library,'' in \emph{Advances in Neural
  Information Processing Systems}, 2019, pp. 8024--8035.

\bibitem{liao2020pgnet}
B.~Liao, C.~Wang, Y.~Wang, Y.~Wang, and J.~Yin, ``Pg-net: Pixel to global
  matching network for visual tracking,'' in \emph{European Conference on
  Computer Vision}.\hskip 1em plus 0.5em minus 0.4em\relax Springer, 2020, pp.
  429--444.

\bibitem{zhang2020ocean}
J.~F. B. L. W.~H. Zhipeng~Zhang, Houwen~Peng, ``Ocean: Object-aware anchor-free
  tracking,'' in \emph{European Conference on Computer Vision (ECCV)}, August
  2020.

\bibitem{fan2017ptav}
H.~Fan and H.~Ling, ``Parallel tracking and verifying: A framework for
  real-time and high accuracy visual tracking,'' in \emph{Proceedings of the
  IEEE International Conference on Computer Vision}, 2017, pp. 5486--5494.

\bibitem{yang2020roam}
T.~Yang, P.~Xu, R.~Hu, H.~Chai, and A.~B. Chan, ``Roam: Recurrently optimizing
  tracking model,'' in \emph{Proceedings of the IEEE/CVF Conference on Computer
  Vision and Pattern Recognition}, 2020, pp. 6718--6727.

\bibitem{zhu2018Dist-RPN}
Z.~Zhu, Q.~Wang, B.~Li, W.~Wu, J.~Yan, and W.~Hu, ``Distractor-aware siamese
  networks for visual object tracking,'' in \emph{Proceedings of the European
  Conference on Computer Vision (ECCV)}, 2018, pp. 101--117.

\bibitem{lee2018MMLT}
H.~Lee, S.~Choi, and C.~Kim, ``A memory model based on the siamese network for
  long-term tracking,'' in \emph{Proceedings of the European Conference on
  Computer Vision (ECCV) Workshops}, 2018.

\bibitem{wang2018describe}
X.~Wang, C.~Li, R.~Yang, T.~Zhang, J.~Tang, and B.~Luo, ``Describe and attend
  to track: Learning natural language guided structural representation and
  visual attention for object tracking,'' \emph{arXiv preprint
  arXiv:1811.10014}, 2018.

\bibitem{guo2017DSiam}
Q.~Guo, W.~Feng, C.~Zhou, R.~Huang, L.~Wan, and S.~Wang, ``Learning dynamic
  siamese network for visual object tracking,'' in \emph{Proceedings of the
  IEEE international conference on computer vision}, 2017, pp. 1763--1771.

\bibitem{bai2020deepDeblur}
Y.~Bai, T.~Xu, B.~Huang, and R.~Yang, ``Deep deblurring correlation filter for
  object tracking,'' \emph{IEEE Access}, vol.~8, pp. 68\,623--68\,637, 2020.

\bibitem{kiani2017nfs}
H.~Kiani~Galoogahi, A.~Fagg, C.~Huang, D.~Ramanan, and S.~Lucey, ``Need for
  speed: A benchmark for higher frame rate object tracking,'' in
  \emph{Proceedings of the IEEE International Conference on Computer Vision},
  2017, pp. 1125--1134.

\bibitem{wang2019deformable}
T.~Wang, H.~Ling, C.~Lang, S.~Feng, and X.~Hou, ``Deformable surface tracking
  by graph matching,'' in \emph{Proceedings of the IEEE/CVF International
  Conference on Computer Vision}, 2019, pp. 901--910.

\bibitem{liu2019deformable}
W.~Liu, Y.~Song, D.~Chen, S.~He, Y.~Yu, T.~Yan, G.~P. Hancke, and R.~W. Lau,
  ``Deformable object tracking with gated fusion,'' \emph{IEEE Transactions on
  Image Processing}, vol.~28, no.~8, pp. 3766--3777, 2019.

\end{thebibliography}
}

\end{document}